%% file: main.tex
\documentclass[10pt]{article}
\usepackage[preprint]{tmlr}
\input{math_commands.tex}

\usepackage{hyperref}
\hypersetup{colorlinks=true, allcolors=blue!50!black}
\usepackage{url}
\usepackage{booktabs}
\usepackage{graphicx}
\usepackage{amsmath}
\usepackage{xcolor}
\usepackage{tikz}
\usetikzlibrary{arrows.meta,positioning,calc,fit,backgrounds,decorations.pathreplacing}
\title{TabOmics: A Small, Open Tabular Foundation Model for Wide Omics Data}

\author{\name Andrea López-García \email andrea.lopez@urjc.es \\
        \addr Department of Signal Theory and Communications, Telematics and Computing\\
      Universidad Rey Juan Carlos, Fuenlabrada, Spain
      \AND
      \name Luis Bote-Curiel \email luis.bote@urjc.es \\
      \addr Department of Signal Theory and Communications, Telematics and Computing\\
      Universidad Rey Juan Carlos, Fuenlabrada, Spain
      \AND
      \name Francisco Javier Gimeno-Blanes \email javier.gimeno@umh.es \\
      \addr Department of Signal Theory and Communications\\
      Universidad Miguel Hernández, Elche, Spain
      \AND
      \name José Luis Rojo-Álvarez \email joseluis.rojo@urjc.es \\
      \addr Department of Signal Theory and Communications, Telematics and Computing\\
      Universidad Rey Juan Carlos, Fuenlabrada, Spain
      \AND
      \name Sergio Muñoz-Romero \email sergio.munoz@urjc.es \\
      \addr Department of Signal Theory and Communications, Telematics and Computing\\
      Universidad Rey Juan Carlos, Fuenlabrada, Spain}

\begin{document}
\maketitle

\begin{abstract}
Omics studies often produce tables with thousands to tens of thousands of molecular features and few samples. Tabular foundation models use diverse strategies to predict labels for such tables, with some relying on in-context learning, while approaches to wide tables may involve subsampling, grouping, or compressing features. We present TabOmics, a small, open family of tabular foundation models with 7.3M parameters, pretrained exclusively on synthetic tables. TabOmics processes all features retained after preprocessing jointly and derives predictions, attention-derived feature-importance scores, and test-sample embeddings from the same forward pass. On 46 classification tasks from three benchmarks of high-dimensional, low-sample-size tables, TabOmics v2.5 with eight ensemble members achieves the highest mean area under the receiver operating characteristic curve (AUROC) among the evaluated models (0.943), outperforming TabPFN-3.5 with eight members on 32 tasks despite having about 30 times fewer parameters. With a single member, TabOmics v2.5 reaches a mean AUROC of 0.941, versus 0.942 for TabPFN-3.5 with eight members, with no significant paired difference under the reported task-level analysis. On synthetic tables, its feature-importance scores achieve feature recovery similar to that of TabPFN-Wide, and its predictions remain above chance at 30,000 features. On breast cancer tasks, the top-ranked genes include PAM50 genes, which define the subtype labels, indicating concordance with the label definition rather than independent biological validation. The code, model weights, training recipes, and experimental artifacts are publicly available.
\end{abstract}

\input{sections/01_introduction}
\input{sections/02_model}
\input{sections/03_synthetic_data}
\input{sections/04_real_data}
\input{sections/05_training}
\input{sections/06_interpretability}
\input{sections/07_evaluation}
\input{sections/08_results}
\input{sections/09_limitations}
\input{sections/10_conclusion}

\bibliography{references}
\bibliographystyle{tmlr}

\appendix
\input{sections/appendix}

\end{document}

%% file: math_commands.tex
\usepackage{amsmath,amsfonts,bm}

\def\eqref#1{equation~\ref{#1}}

\def\1{\bm{1}}

\DeclareMathAlphabet{\mathsfit}{\encodingdefault}{\sfdefault}{m}{sl}
\SetMathAlphabet{\mathsfit}{bold}{\encodingdefault}{\sfdefault}{bx}{n}



%% file: sections/01_introduction.tex
\section{Introduction}
\label{sec:introduction}

Omics technologies, including genomics, transcriptomics, epigenomics, proteomics, and metabolomics, characterize biological samples by generating high-dimensional profiles at bulk or single-cell resolution. Paired with clinical or phenotypic labels, such as disease status, tumor subtype, or cellular state, these profiles can train predictive models that link molecular features to the labels \citep{hasin2017multiomics}.

Bulk omics studies usually yield data matrices with few samples and thousands to tens of thousands of molecular features \citep{clarke2008properties}. This high-dimensional, low-sample-size (HDLSS) regime increases the risk of overfitting and makes it difficult to learn relationships that generalize to new samples. Classical machine learning pipelines address this problem through regularization, feature selection, or feature extraction \citep{guyon2003introduction,clarke2008properties}. Feature selection keeps a subset of the original features, whereas feature extraction maps them into a lower-dimensional representation. Both can improve generalization, but reducing or transforming the feature space may discard predictive information, and strong regularization can limit the relationships the model captures.

Learning from very few samples is itself a bottleneck. One approach is to reuse knowledge acquired across other tasks instead of fitting each new dataset from scratch. Tabular foundation models (TFMs), which operate on tabular datasets (or tables), follow this approach: they are pretrained across diverse datasets or tasks, which can reduce the amount of task-specific data needed for prediction \citep{zhu2023xtab,hollmann2025tabpfn}. Most use Transformer architectures \citep{vaswani2017attention}, whose attention mechanisms capture relationships within the inputs. Current approaches use pretraining in three ways. First, cross-table methods, such as XTab \citep{zhu2023xtab} and CARTE \citep{kim2024carte}, learn transferable representations for heterogeneous tables through self-supervised reconstruction objectives or contrastive learning, and are then fine-tuned on the downstream dataset. Second, large language model (LLM) approaches, such as TabLLM \citep{hegselmann2023tabllm}, serialize tabular inputs as text and use pretrained language models through prompting or task-specific fine-tuning. Third, prior-data fitted networks (PFNs) learn from synthetic prediction tasks sampled from a specified prior to approximate Bayesian posterior predictive inference \citep{muller2022transformers}. TabPFN \citep{hollmann2023tabpfn} uses this approach to perform in-context learning (ICL), predicting labels for unseen tables from labeled examples provided as context, without task-specific parameter updates. Its successor, TabPFNv2 \citep{hollmann2025tabpfn}, together with TabICL \citep{qu2025tabicl} and TabFM \citep{kong2026tabfm}, follows the same synthetic-pretraining and ICL paradigm.

However, learning from few samples does not resolve the problem of high dimensionality. Earlier Transformer-based TFMs faced computational and generalization limitations on extremely wide tables \citep{hollmann2025tabpfn,ye2025closerlook,kolberg2025tabpfnwide}. Full self-attention has a computational cost that grows quadratically with the number of features, which increases memory requirements and inference time \citep{vaswani2017attention,kolberg2025tabpfnwide}. Generalization is also affected, particularly in bulk omics, where noisy or irrelevant measurements can obscure the biological signals associated with the labels \citep{clarke2008properties}. These difficulties may reflect insufficient exposure to wide tables during pretraining rather than a fundamental architectural limitation. TabPFN-Wide supports this interpretation: continued pretraining on synthetic HDLSS data, generated with a feature-widening prior that incorporates noise and inter-feature dependencies, reduced the generalization gap \citep{kolberg2025tabpfnwide}. Other recent approaches pair suitable pretraining priors with feature subsampling, grouping, internal chunking, memory-efficient attention, or compression into dense row representations \citep{ye2025closerlook,qu2026tabiclv2,jaeger2026tabpfn35}. TabPFN-3.5, for instance, accepts up to 20,000 columns \citep{priorlabs2026models}, although each ensemble member processes only a subset of them (Section~\ref{sec:models}), and TabICLv2 has shown strong performance on benchmarks with up to 2,000 features \citep{tabicl2026documentation}.

Prediction alone is not enough for many omics applications, which also require identifying the molecular features that drive the predictions. These features are candidates for biomarker discovery and subsequent biological validation, which connects predictive modeling with scientific discovery \citep{ditz2023comic} and makes interpretability an important component of predictive omics analysis. Post hoc attribution methods, including Shapley-value explanations, have been applied to the original TabPFN \citep{rundel2024interpretable}, but perturbation-based approaches can become computationally expensive with thousands of features \citep{lundberg2017unified,kolberg2025tabpfnwide}. Complementary approaches investigate internal model components through mechanistic analyses or attention-based feature-importance scores. \citet{bilos2026mechanistic} examined how different TFMs use labeled context samples and identified distinct approximate prediction rules, supported by interventions on internal model components. More directly addressing molecular feature relevance, TabPFN-Wide derived feature-importance scores from internal attention maps. Its highly ranked features were more predictive and overlapped with previous biological findings in cancer datasets \citep{kolberg2025tabpfnwide}. Sample embeddings extracted from TFMs offer a complementary view of the relationships among samples and can support downstream predictive tasks, as \citet{ye2025closerlook} explored in TabPFNv2.

We present TabOmics, a small, open TFM family that combines wide-table prediction, feature importance, and sample representations within a single model. The code, model weights, and staged pretraining recipes, which can be rerun from scratch on a single graphics processing unit (GPU), are available in the \texttt{tabomics} package\footnote{\url{https://github.com/lbotecur/tabomics}}. TabOmics builds on nanoTabPFN \citep{pfefferle2025nanotabpfn,nanotabpfnrepo}, a lightweight reimplementation of TabPFNv2, and on its implementation in TFM-Playground \citep{tfmplayground}. It combines the synthetic-data prior of TabICLv2 \citep{qu2026tabiclv2} with a reimplementation of the feature-widening procedure used by TabPFN-Wide \citep{kolberg2025tabpfnwide,tabpfnwiderepo}. As in TabPFNv2 \citep{hollmann2025tabpfn}, nanoTabPFN is extended with missing-value indicators, random column embeddings, and a test-row indicator in the label column.

We release two classification models, TabOmics v2 and v2.5, pretrained exclusively on synthetic tables in stages: training on narrow tables first, then a stage in which a fraction of the tables are widened, and a final stage at a lower learning rate. At inference, TabOmics processes all features retained after preprocessing jointly and predicts test labels through ICL without task-specific parameter updates. TabOmics also supports optional fine-tuning \citep{tabicl2026software,rubachev2025finetuning}, which is not evaluated in this work. Joint processing is feasible because attention is computed without materializing the full attention matrices, so the memory required for feature-wise attention grows linearly with the number of features, although its computational cost remains quadratic. Following TabPFN-Wide, TabOmics derives feature importance from attention, but it averages the scores over the test samples only, so that they reflect attention from the samples being predicted. Following \citet{ye2025closerlook}, TabOmics also provides sample embeddings: test samples are embedded using all training samples as context, whereas training samples are embedded out of fold, so that no embedding ever uses its own label.

As in the evaluation of TabPFN-Wide, both models are assessed in terms of predictive performance and computational cost across 46 classification tasks from three HDLSS benchmarks, two of which focus on multi-omics cancer data. The feature-importance scores are assessed on synthetic datasets with known informative features and on cancer transcriptomic data, where the predictive value and stability of the resulting gene rankings and their concordance with established subtype markers are examined. The sample embeddings are also evaluated as inputs to simple classifiers and visualized on a breast cancer task. With eight ensemble members, that is, forward passes on randomly permuted and relabeled copies of the table, TabOmics v2.5 achieves the highest mean area under the receiver operating characteristic curve (AUROC) among the models compared. With a single member, it performs comparably to TabPFN-3.5 with eight ensemble members, a model with about 30 times as many parameters.

Sections~\ref{sec:model}--\ref{sec:interpretability} describe the model, synthetic- and real-data preparation, training, feature importance, and sample embeddings. Sections~\ref{sec:evaluation} and~\ref{sec:results} present the evaluation protocol and results, respectively. Section~\ref{sec:limitations} discusses the findings and limitations, and Section~\ref{sec:conclusion} concludes.

%% file: sections/02_model.tex
\section{Model}
\label{sec:model}

TabOmics is a Transformer that predicts the labels of the test rows of a table from its labeled training rows in a single forward pass, without updating its parameters. This section describes its architecture, how a table is encoded as its input, how it is run at inference, and how it can optionally be fine-tuned.

\subsection{Architecture}
\label{sec:architecture}

TabOmics uses the nanoTabPFN architecture \citep{pfefferle2025nanotabpfn,nanotabpfnrepo}, a compact reimplementation of the TabPFNv2 architecture \citep{hollmann2025tabpfn}, as implemented in TFM-Playground \citep{tfmplayground}. A table with $N$ training rows, $M$ test rows to predict, and $C$ features, where each row corresponds to a sample, is represented as a grid of tokens, one per table cell. Here, training rows denote the labeled context used for prediction, not rows used to update the model parameters. Each feature token is an $E$-dimensional vector that encodes the numerical value or category code of a cell, together with its missing-value status and column position. A label column is appended, which gives $C+1$ tokens per row, and each label token encodes a class index and a flag that indicates whether the row is a training or a test row. For a training row, the class index is its class, whereas for a test row, it is replaced by a placeholder (Section~\ref{sec:encoding}).

The model stacks $L$ Transformer blocks. Each block applies three steps in turn, and each step $f$ is followed by a residual connection and layer normalization \citep[post-norm;][]{vaswani2017attention}, $\mathbf{x} \leftarrow \mathrm{LayerNorm}(\mathbf{x} + f(\mathbf{x}))$, where $\mathbf{x}$ denotes the current token representations:
\begin{enumerate}
  \item \emph{Attention between features}: within each row, every token attends to all the tokens of that row, including the label token.
  \item \emph{Attention between rows}: within each column, every token attends to the tokens of the training rows only, so that test rows never influence each other or the training rows.
  \item MLP with Gaussian error linear unit (GELU) activation \citep{hendrycks2016gelu}, applied to every token.
\end{enumerate}

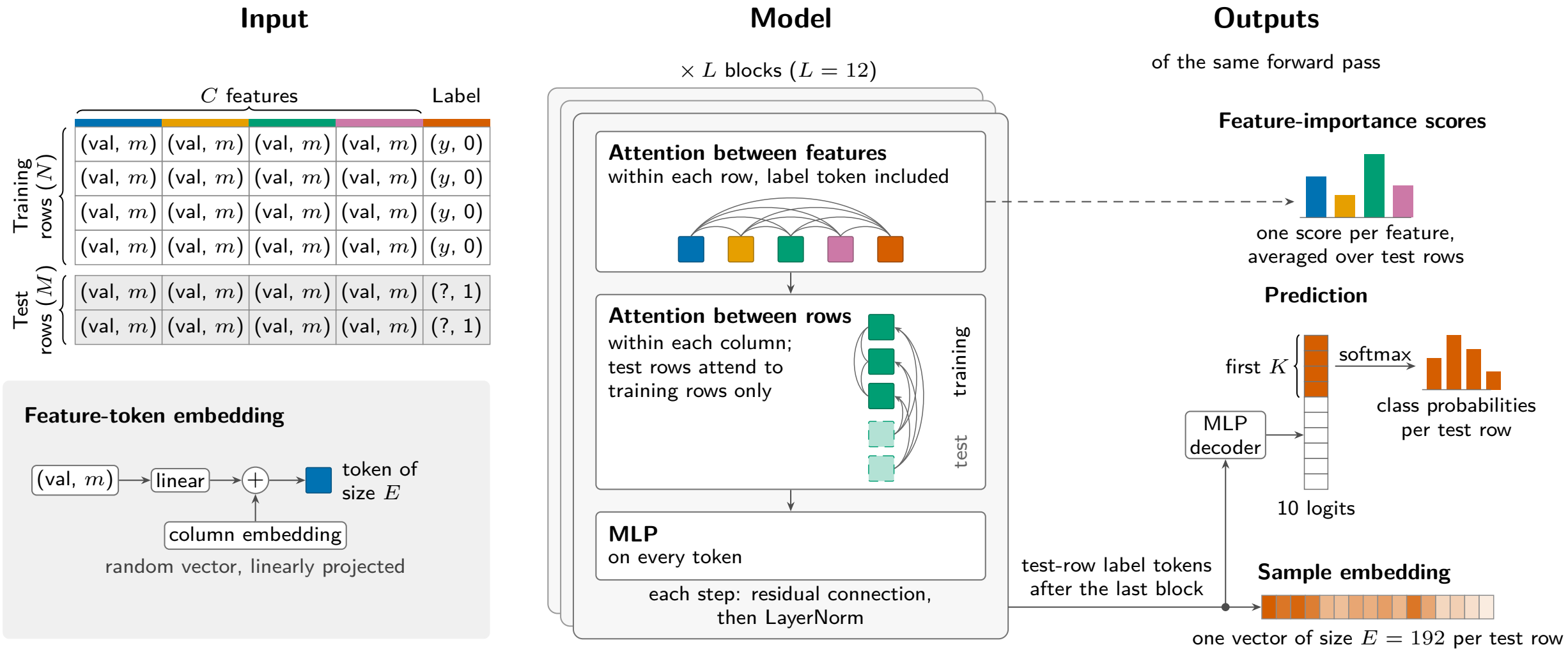
\begin{figure}[t]
  \centering
  \resizebox{\linewidth}{!}{\input{figures/architecture}}
  \caption{Architecture of TabOmics. An input table is represented as a grid of feature and label tokens for $N$ training rows and $M$ test rows. Each of the $L=12$ Transformer blocks applies feature-wise attention within rows, row-wise attention within columns, and a token-wise two-layer multilayer perceptron (MLP). During row-wise attention, test tokens attend only to the training rows. The final test-row label tokens provide class probabilities and 192-dimensional sample embeddings, while label-to-feature attention provides one score per retained feature. Solid and pale tokens denote training and test rows, respectively. Colors identify columns.}
  \label{fig:architecture}
\end{figure}

After the last block, a two-layer MLP with GELU activation maps the label token of each test row to 10 class logits. For a task with $K$ classes, the first $K$ logits are retained and converted to class probabilities with a softmax. Both TabOmics models use $L=12$ blocks, $E=192$, six attention heads, and an MLP hidden size of 768, for a total of 7.3M parameters. Figure~\ref{fig:architecture} shows the architecture.

The architecture also supports regression, as in TabPFNv2 \citep{hollmann2025tabpfn}, by decoding the label token into a distribution over bins of the standardized label \citep{muller2022transformers}, but no regression model is released.

\subsection{Input encoding}
\label{sec:encoding}

\paragraph{Preprocessing.} At inference, an input table is converted into numerical inputs before it is passed to the model, using a preprocessing that is fitted on the training rows and applied unchanged to the test rows. Features that are constant or entirely missing in the training rows are dropped, and categorical columns, which can be specified explicitly or identified automatically from the training rows, are encoded as follows: a categorical feature $j$ with $K_j$ categories observed in the training rows receives integer codes $0, \dots, K_j-1$, while categories that appear only in the test rows are treated as missing values. Training class labels are assumed to have no missing values and are mapped to indices $0, \dots, K-1$, where $K \le 10$ is the number of classes. This preprocessing is not applied during pretraining, where tables are assumed to already have the resulting format: numerical features, with categorical features and class labels represented by integer codes.

\paragraph{Features.} Inside the model, each feature is standardized, during both pretraining and inference, using the mean and sample standard deviation of its observed values in the training rows, so that test rows never contribute to these statistics. Categorical codes are standardized in the same way, so the model receives only numbers and no column-type information. Standardized values are then clipped to $[-c, c]$, with $c=100$ during pretraining and $c=10$ at inference (Section~\ref{sec:clip}). Following TabPFNv2 \citep{hollmann2025tabpfn}, missing values are replaced by zero after standardization, which corresponds to the mean of the observed training values, and are flagged by a binary indicator. Each cell is thus represented as a (value, missing-value indicator) pair and embedded by a linear layer shared across all feature cells.

\paragraph{Feature identity.} Because this linear layer is shared across columns, cells with identical inputs receive the same embedding, whatever their column, and attention between features does not use the order of the tokens either: it is permutation equivariant, since reordering the input tokens only reorders the output tokens in the same way. Without further information, two columns with the same values would therefore be indistinguishable, so the model needs an explicit column identifier. Following TabPFNv2 \citep{hollmann2025tabpfn}, each feature-column position receives a random vector of dimension $E/4$, drawn from a standard normal distribution with a fixed seed, so that each position always receives the same vector. A learned linear layer then projects this vector to dimension $E$, and the result is added to every cell token in the column. These identifiers distinguish columns but carry no information about their semantic roles, since the model is pretrained on tables whose columns are in random order. The model is therefore only approximately invariant to the order of the columns, and the ensemble members, which reorder the columns at random, reduce this dependence (Section~\ref{sec:inference}).

\paragraph{Labels.} Each label token embeds a pair formed by a class index and a test-row indicator, with a linear layer shared by all rows, as in TabPFNv2 \citep{hollmann2025tabpfn}. For a training row, the pair is its class index and 0, whereas for a test row, whose class the model predicts, the class index is replaced by a placeholder, the mean of the class indices of the training rows of the table, and the indicator is 1. The indicator is needed because, without it, a placeholder equal to a class index would give the label token of a test row the same embedding as that of a training row of the corresponding class.

\subsection{Inference}
\label{sec:inference}

\paragraph{In-context prediction.} Given a labeled training set and a set of test rows, both preprocessed and encoded as in Section~\ref{sec:encoding}, TabOmics predicts the test rows in a single forward pass with no parameter update. Inference can run in single precision or, as in pretraining, in bfloat16 mixed precision, with the class logits converted to single precision before the softmax. All the features retained by the preprocessing are processed jointly, without feature subsampling, grouping, or compression. Optionally, as in TabPFN and TabICL, predictions can be averaged over an ensemble, in which each member is a forward pass of the same model, with the same pretrained weights, on a transformed copy of the table. The first member uses the preprocessed table as it is, whereas each further member applies a random permutation to the order of the columns, to the category codes of each categorical feature, and to the class indices, identically for the training and the test rows. The prediction should be invariant to these transformations, so the class probabilities of each member are mapped back to the original classes and averaged, which reduces the sensitivity of the prediction to them. The number of members is set by the parameter \texttt{n\_estimators} of the \texttt{tabomics} package.

\paragraph{Feature clipping.}
\label{sec:clip}
Clipping bounds the standardized values, which can be arbitrarily large when a feature varies little in the training rows: its standard deviation is then close to zero, and a value away from the training mean is divided by it. The models were pretrained with $c=100$, the clipping threshold of the nanoTabPFN implementation used as the starting point, and this threshold is reduced to $c=10$ for inference and stored in the released model configuration. In the synthetic training tables, whose outliers the prior clamps before standardizing them (Section~\ref{sec:graph_prior}), standardized values beyond 10 in absolute value are rare, but sparse real features can produce much larger ones. For a binary feature that takes the value 1 in a fraction $\phi$ of the training rows and 0 otherwise, the standardized value of 1 is approximately $\sqrt{(1-\phi)/\phi}$, about 30 for $\phi = 0.1\%$. Clipping at 10 therefore bounds such values to the range the model saw most often in pretraining, and its effect is quantified in Section~\ref{sec:results_ablations}.

\paragraph{Memory-efficient attention.} Attention between features and attention between rows are computed with the \texttt{scaled\_dot\_product\_attention} function of PyTorch \citep{paszke2019pytorch}, in pretraining and at inference. This function selects its implementation according to the device, data type, and input shapes, and the selection may change between PyTorch versions. In the CUDA environment specified in the \hyperref[sec:hardware_software]{Hardware and software environment} section, this function uses fused kernels such as FlashAttention \citep{dao2022flashattention,dao2024flashattention2} and memory-efficient attention \citep{rabe2021selfattention}. Unlike the nanoTabPFN implementation used as the starting point, which explicitly materialized the full attention matrices, these kernels compute attention in blocks without storing all the weights simultaneously. As a result, the memory required for attention between features grows linearly with the number of features, although the computational cost remains quadratic. This agrees with the measured memory usage: on the widest task, with 57,133 features, storing the attention weights of a single row across all six heads would require about 39\,GB in bfloat16, more than the peak of 27.7\,GB measured for TabOmics in the reported hardware environment (Section~\ref{sec:results_cost}). When feature importance is requested (Section~\ref{sec:interpretability}), only the label-to-feature attention weights are computed explicitly, so this additional step also requires memory linear in the number of features.

\paragraph{Memory chunks and low-memory mode.} Once the attention weights are no longer materialized, the memory of a forward pass on wide tables is dominated by the token grid, which holds $(N+M)\times(C+1)\times E$ values, and two options reduce the peak memory at inference. First, each step of a block can be applied in chunks (eight by default) and each result written in place: attention between features in chunks of rows, attention between rows in chunks of columns, and the MLP in chunks of tokens, so that every attention still sees its full context. Chunking reduces the intermediate tensors of each step, but not the grid itself. Second, in low-memory mode, layer normalization is applied within each chunk and activations are kept in the precision of the input (bfloat16 under mixed precision), instead of materializing full-grid tensors in single precision.
This mode computes the same function as the standard path, subject to rounding differences under mixed precision. All reported results use the low-memory path. 

\subsection{Fine-tuning}
\label{sec:finetuning}

Although TabOmics is designed for in-context prediction, the \texttt{tabomics} package can also fine-tune it on one or more tables, following the \texttt{tabicl} package \citep{tabicl2026software} and \citet{rubachev2025finetuning}: a copy of the model is trained on episodes drawn from the training rows, each divided into a labeled context and query rows whose labels provide the training objective. Fine-tuning is not evaluated in this paper.

%% file: figures/architecture.tex
\definecolor{archA}{HTML}{0072B2}\definecolor{archB}{HTML}{E69F00}\definecolor{archC}{HTML}{009E73}
\definecolor{archD}{HTML}{CC79A7}\definecolor{archY}{HTML}{D55E00}
\begin{tikzpicture}[x=1mm, y=1mm, >={Stealth[length=3pt, width=2.4pt]}, line width=0.4pt,
  every node/.style={font=\sffamily\fontsize{6.5}{7.5}\selectfont, inner sep=0.8pt},
  head/.style={anchor=base, font=\sffamily\fontsize{9}{10}\selectfont\bfseries},
  title/.style={font=\sffamily\fontsize{7}{8}\selectfont\bfseries},
  cell/.style={draw=black!45, line width=0.3pt, minimum height=3.8mm, inner sep=0},
  tok/.style={draw=#1!70!black, fill=#1, line width=0.3pt, minimum size=2.8mm, inner sep=0, rounded corners=0.5pt},
  ttok/.style={draw=#1, fill=#1!30, densely dashed, line width=0.3pt, minimum size=2.8mm, inner sep=0,
               rounded corners=0.5pt},
  box/.style={draw=black!55, fill=white, rounded corners=1.5pt, inner sep=1.4pt, align=center},
  stage/.style={draw=black!55, fill=white, rounded corners=1.5pt},
  arr/.style={->, black!70},
]
\node[head] at (31, 1.5) {Input};
\foreach \c/\col in {0/archA, 1/archB, 2/archC, 3/archD} {
  \fill[\col] (9+\c*9.6, -9.5) rectangle ++(9.6, 0.9);
  \foreach \r in {0,...,3} { \node[cell, minimum width=9.6mm] at (13.8+\c*9.6, -11.4-\r*3.8) {(val, $m$)}; }
  \foreach \r in {0,1} { \node[cell, minimum width=9.6mm, fill=black!8] at (13.8+\c*9.6, -27.8-\r*3.8) {(val, $m$)}; }
}
\fill[archY] (47.4, -9.5) rectangle ++(7.4, 0.9);
\foreach \r in {0,...,3} { \node[cell, minimum width=7.4mm] at (51.1, -11.4-\r*3.8) {($y$, 0)}; }
\foreach \r in {0,1} { \node[cell, minimum width=7.4mm, fill=black!8] at (51.1, -27.8-\r*3.8) {(?, 1)}; }
\draw[decorate, decoration={brace, amplitude=2pt}] (9.2, -8) -- (47.2, -8) node[midway, above=2.5pt] {$C$ features};
\node[above=2.5pt] at (51.1, -8) {Label};
\draw[decorate, decoration={brace, amplitude=2pt, mirror}] (8, -9.6) -- (8, -24.7)
  node[midway, left=2pt, rotate=90, anchor=south, align=center] {Training\\rows ($N$)};
\draw[decorate, decoration={brace, amplitude=2pt, mirror}] (8, -25.9) -- (8, -33.5)
  node[midway, left=2pt, rotate=90, anchor=south, align=center] {Test\\rows ($M$)};
\fill[black!6, rounded corners=2pt] (1, -37.5) rectangle (54.8, -66);
\node[title, anchor=west] at (3, -41.5) {Feature-token embedding};
\node[box] (pair) at (9, -48.5) {(val, $m$)};
\node[box, right=3.5 of pair] (lin) {linear};
\node[draw=black!55, fill=white, circle, inner sep=0.3pt, right=3.5 of lin] (plus) {$+$};
\node[tok=archA, right=4 of plus] (token) {};
\node[right=0.8 of token, align=left] {token of\\size $E$};
\node[box, below=3 of plus] (col) {column embedding};
\node[below=0.6 of col, text=black!75] {random vector, linearly projected};
\draw[arr] (pair) -- (lin); \draw[arr] (lin) -- (plus); \draw[arr] (col) -- (plus); \draw[arr] (plus) -- (token);

\node[head] at (88, 1.5) {Model};
\node at (86.6, -3.4) {$\times\,L$ blocks ($L = 12$)};
\foreach \d in {2,1} {
  \draw[draw=black!40, fill=black!3, rounded corners=3pt] (64-\d*1.4, -8+\d*1.4) rectangle (112-\d*1.4, -66+\d*1.4);
}
\draw[draw=black!55, fill=black!3, rounded corners=3pt] (64, -8) rectangle (112, -66);
\node[stage, minimum width=43mm, minimum height=15.5mm] (fa) at (88, -17.75) {};
\node[title, anchor=west] at (67.5, -12.3) {Attention between features};
\node[anchor=west] at (67.5, -15) {within each row, label token included};
\foreach \i/\col in {0/archA, 1/archB, 2/archC, 3/archD, 4/archY} { \node[tok=\col] (fa\i) at (77+\i*5.5, -23) {}; }
\foreach \i in {0,...,3} {
  \pgfmathtruncatemacro{\n}{\i+1}
  \foreach \j in {\n,...,4} { \draw[black!60, line width=0.3pt] (fa\i.north) to[out=55, in=125, looseness=0.8] (fa\j.north); }
}
\node[stage, minimum width=43mm, minimum height=21.5mm] (ra) at (88, -38.75) {};
\node[title, anchor=west] at (67.5, -30.3) {Attention between rows};
\node[anchor=north west, text width=23mm] at (67.5, -32.1) {within each column; test rows attend to training rows only};
\foreach \i/\y in {0/-31.6, 1/-35.4, 2/-39.2} { \node[tok=archC] (tr\i) at (98, \y) {}; }
\foreach \i/\y in {0/-43.4, 1/-47.2} { \node[ttok=archC] (te\i) at (98, \y) {}; }
\foreach \i/\j in {0/1, 1/2, 0/2} { \draw[black!60, line width=0.3pt] (tr\j.west) to[out=150, in=210, looseness=0.8] (tr\i.west); }
\foreach \t in {0,1} { \foreach \i in {0,1,2} {
  \draw[->, >={Stealth[length=2.2pt, width=1.8pt]}, black!60, line width=0.3pt] (te\t.east) to[out=30, in=-30, looseness=0.8] (tr\i.east);
} }
\node[rotate=90] at (106.8, -35.4) {training};
\node[rotate=90, text=black!60] at (106.8, -45.3) {test};
\node[stage, minimum width=43mm, minimum height=7.5mm] (mlp) at (88, -55.75) {};
\node[title, anchor=west] at (67.5, -54.4) {MLP};
\node[anchor=west] at (67.5, -57.1) {on every token};
\draw[arr] (fa) -- (ra); \draw[arr] (ra) -- (mlp);
\node[align=center] at (88, -62.5) {each step: residual connection,\\then LayerNorm};

\node[head] at (140.5, 1.5) {Outputs};
\node at (140.5, -2.5) {of the same forward pass};
\node[title] at (150, -9) {Feature-importance scores};
\foreach \i/\h/\col in {0/4.5/archA, 1/2.5/archB, 2/7/archC, 3/3.5/archD} { \fill[\col] (144.9+\i*3.2, -19.5) rectangle ++(2.2, \h); }
\draw[black!55, line width=0.3pt] (144.2, -19.5) -- (156.5, -19.5);
\node[align=center] at (150.4, -22.6) {one score per feature,\\averaged over test rows};
\draw[->, densely dashed, black!70] (fa.east) -- (143.6, -17.75);
\node[title] at (146, -28) {Prediction};
\node[box] (dec) at (136, -43.5) {MLP\\decoder};
\foreach \i in {0,...,9} {
  \pgfmathsetmacro{\f}{\i<4 ? 100 : 0}
  \node[draw=black!55, line width=0.3pt, fill=archY!\f, minimum width=2.6mm, minimum height=1.7mm, inner sep=0]
    (lg\i) at (146, -33.35-\i*1.7) {};
}
\draw[decorate, decoration={brace, amplitude=1.5pt}] ($(lg3.south west)-(0.5,0)$) -- ($(lg0.north west)-(0.5,0)$)
  node[midway, left=2pt] {first $K$};
\node[below=0.8 of lg9] {10 logits};
\draw[arr] (dec) -- (dec -| lg0.west);
\draw[arr] (148.1, -35.9) -- (157, -35.9) node[midway, above=0.8pt, text=black] {softmax};
\foreach \i/\h in {0/3.5, 1/6, 2/4.5, 3/2} { \fill[archY] (158.2+\i*2.2, -38.5) rectangle ++(1.6, \h); }
\draw[black!55, line width=0.3pt] (157.7, -38.5) -- (165.3, -38.5);
\node[align=center] at (161.5, -41.7) {class probabilities\\per test row};
\node[title, anchor=west] at (139.2, -58.8) {Sample embedding};
\foreach \i/\f in {0/100, 1/85, 2/95, 3/80, 4/45, 5/40, 6/55, 7/50, 8/60, 9/45, 10/85, 11/55, 12/25, 13/30, 14/20, 15/12} {
  \node[draw=black!45, line width=0.3pt, fill=archY!\f, minimum width=1.6mm, minimum height=2.6mm, inner sep=0]
    (em\i) at (140.8+\i*1.6, -62.5) {};
}
\node[below=0.8 of em7.south east] {one vector of size $E = 192$ per test row};
\draw[black!70] (112, -62.5) -- (136, -62.5);
\draw[arr] (136, -62.5) -- (dec.south);
\draw[arr] (136, -62.5) -- (em0.west);
\fill[black!70] (136, -62.5) circle (0.45);
\node[align=center] at (124, -59.1) {test-row label tokens\\after the last block};
\end{tikzpicture}

%% file: sections/03_synthetic_data.tex
\section{Synthetic data}
\label{sec:prior}

TabOmics is pretrained only on synthetic classification tables drawn from a prior, a distribution over tables defined by a random generator. The synthetic-data pipeline combines the graph-based prior of TabICLv2 with randomly introduced missing values and a feature-widening procedure used during the later training stages.

\subsection{Graph prior}
\label{sec:graph_prior}

Each table is drawn from a random structural causal model, using the TabICLv2 version \citep{qu2026tabiclv2} of the prior introduced by TabPFN \citep{hollmann2023tabpfn}. The model is defined by a random directed acyclic graph in which root nodes receive random input vectors, drawn independently for each row, and every other node applies a random function to the outputs of its parents. The graph and its functions are shared across all rows of the table. The features and the label are extracted from the outputs of randomly assigned nodes through conversion mechanisms such as discretization into categories. The graph can thus create causal dependencies between the features and the label, as well as shared causes among them. Graphs in which the label shares no ancestor with any feature are resampled. Features may be numerical or categorical, with categorical features initially encoded as integers with fewer than 10 categories.

The generator samples the number of rows (log-uniformly), the number of features (approximately uniformly), the number of classes (uniformly from 2 to 10), the proportion of training rows (uniformly between 30\% and 90\%), and the proportion of categorical features together with the number of categories of each. Features that are constant across all rows are dropped, such that a table can retain fewer features than initially sampled. In addition, a split is accepted only if the training and test rows contain the same set of at least two classes. The ranges for the numbers of features and rows depend on the training stage and are given in Table~\ref{tab:stages}.

The released models were pretrained on tables produced by version 2.2.0 of the \texttt{tabicl} package \citep{tabicl2026software}, which implements TabICLv2. In this generator, outliers are clamped and every feature, including categorical codes, is standardized using all rows of the synthetic table. The model subsequently applies its own standardization using only the training rows. Using all rows in the generator is part of constructing a synthetic task, whereas on real tables the normalization statistics used by the model are computed from the training rows only. The generator is included in the \texttt{tabomics} package, which documents its changes. Random seeding and the reproducibility of the released checkpoints are discussed in Section~\ref{sec:reproducibility}.

\subsection{Feature widening}
\label{sec:widening}

The prior generates tables with at most a few hundred features, whereas omics tables can contain thousands of correlated, noisy, or task-irrelevant features. Following TabPFN-Wide \citep{kolberg2025tabpfnwide}, the synthetic tables are widened in the later stages of pretraining by generating new features from the existing ones. For each widened batch, the number of new features $A \sim \mathcal{U}\{200, \dots, 3000\}$, an inclusion probability $q \sim \mathcal{U}[0, 0.05]$, and a noise scale $\sigma \sim \mathcal{U}[0, 1]$ are sampled. The $C$ original features are retained with probability 0.5. Although these choices are shared across all tables in the batch, the new features are generated separately for each table, preserving its proportion of continuous and categorical features. In the runs that produced the released models, features with at most 20 distinct values were treated as categorical, following the rule of TabPFN-Wide.

\begin{itemize}
\item \emph{Continuous features} (Algorithm~1 of TabPFN-Wide): each new feature is a linear combination of the continuous features of the table, each included independently with probability $q$ and assigned a standard normal weight. Gaussian noise is added with standard deviation $\sigma$ times that of the combination. If the combination has zero variance, for example because no feature was included, the noise has standard deviation $\sigma$, thereby producing a pure-noise feature.
\item \emph{Categorical features} (Algorithm~2 of TabPFN-Wide): for each new feature, $d = \max(1, \lfloor q\,C_{\mathrm{cat}} \rfloor)$ of the $C_{\mathrm{cat}}$ categorical features of the table are selected as donors. In each row, the new feature copies the value of a randomly selected donor. Its number of distinct values is then capped at a number between 3 and 20, drawn from a discrete exponential distribution. If this limit is exceeded, the most frequent categories are retained, and each remaining category is merged into a randomly selected retained category.
\end{itemize}

If the original features are retained, they are shuffled among the new ones, giving $C+A$ features. Otherwise, only the $A$ new features remain. In the widening stages, the prior generates tables with up to 350 features, so a widened table can contain up to 3,350 features. Each batch is widened with probability 0.3, and the remaining batches are left unwidened so that the model continues to see narrow tables. 

\subsection{Missing values}
\label{sec:missing}

After feature widening, whether or not widening is applied to a given batch, missing values are introduced into the generated tables generated tables throughout pretraining. For each batch, a rate $\rho \sim \mathcal{U}[0, 0.1]$ is sampled, and each feature value is independently set to missing with probability $\rho$ (missing completely at random). Labels are never missing. This exposes the model to missing inputs and their indicators (Section~\ref{sec:encoding}).

%% file: sections/04_real_data.tex
\section{Real data}
\label{sec:real_data}

TabOmics is evaluated on three benchmarks of HDLSS classification tables, measuring predictive performance and computational cost. This section describes the benchmarks, summarized in Table~\ref{tab:benchmarks_description}.

\subsection{Benchmarks}
\label{sec:benchmarks}

The three benchmarks follow the preparation procedures in the code of TabPFN-Wide \citep{kolberg2025tabpfnwide, tabpfnwiderepo}, with the modifications described below:
\begin{itemize}
  \item \emph{HDLSS}, as named by TabPFN-Wide, comprises the 15 high-dimensional datasets selected from the scikit-feature repository \citep{li2017featureselection}, containing between 500 and 22,283 features across diverse domains (gene expression, text, image data, and one synthetic dataset). Specifically, the gene expression and biological datasets include ALLAML, for acute lymphoblastic versus acute myeloid leukemia, and leukemia. Both contain the same patient samples and originate from the same acute-leukemia gene-expression measurements, but are provided as separate benchmark files, one with continuous expression values and the other with a discretized transformation. They also include CLL\_SUB\_111 for chronic lymphocytic leukemia subtypes, GLI\_85 for glioma tumor classification, Prostate\_GE for prostate cancer gene expression, SMK\_CAN\_187 for smoking-related cancer classification, TOX\_171 for toxicological classification, colon for colon cancer tissue gene expression, and lung for multiclass lung cancer classification, alongside text and image datasets (BASEHOCK, PCMAC, RELATHE, arcene, gisette, and madelon). The folds of TabPFN-Wide are used: stratified 3-fold cross-validation repeated 10 times, or 3 times for datasets with at least 2,500 rows.
  \item \emph{MLOmics} comprises five cancer types from The Cancer Genome Atlas (TCGA), as provided by MLOmics \citep{yang2025mlomics}: breast invasive carcinoma (BRCA), colon adenocarcinoma (COAD), glioblastoma multiforme (GBM), brain lower-grade glioma (LGG), and ovarian serous cystadenocarcinoma (OV). Each cancer has tasks for messenger RNA (mRNA) expression, copy number variation (CNV), deoxyribonucleic acid (DNA) methylation, microRNA (miRNA), and the four omics combined, giving 25 tasks. Stratified 5-fold splits generated with fixed seeds are used. For the mRNA tasks, the ordered test-fold labels match those published by TabPFN-Wide. Exact fold assignments were not published for the other modalities.
  \item \emph{Shamir} comprises breast, glioblastoma, and sarcoma datasets from the multi-omics cancer benchmark of \citet{rappoport2018multiomic}. Each has tasks for mRNA expression, DNA methylation, miRNA, and the three omics combined, giving 12 tasks. Stratified 5-fold cross-validation is used but, unlike in TabPFN-Wide, classes with fewer patients than folds are excluded and the samples are sorted before splitting. Sorting makes the folds reproducible, since the sample order of TabPFN-Wide, and therefore its folds, changes between runs.
\end{itemize}

\begin{table}[t]
  \caption{Summary of the 46 classification tasks used to evaluate TabOmics. Each HDLSS row represents one dataset, whereas each MLOmics and Shamir row groups the modality-specific tasks of one cancer type. The modalities included in each task group are listed below its name. The table reports the numbers of samples, input features, classes, the size of the smallest class, and the cross-validation scheme. For MLOmics and Shamir, feature counts are ranges across the listed modalities. Fold-specific preprocessing may further reduce the number of retained features. CV denotes the cross-validation scheme: $10\times3$ indicates 10 repetitions of stratified 3-fold cross-validation, whereas 5 indicates stratified 5-fold cross-validation. The four HDLSS development datasets and the excluded LGG methylation and combined tasks are not included.}
  \label{tab:benchmarks_description}
  \centering
  \small
  \setlength{\tabcolsep}{4pt}
  \input{tables/benchmarks_description}
\end{table}

These benchmarks facilitate comparison with TabPFN-Wide. All folds are generated with fixed seeds and saved alongside the corresponding tasks in the experiment repository.

No transformation is fitted on the full set of samples. Features with duplicated names and samples with duplicated identifiers are removed when the data are read. In MLOmics, features that are zero in every training row are dropped, and missing values are imputed with training-row means within each fold, whereas TabPFN-Wide performs both operations over all samples. Likewise, in Shamir, the standardization over all samples applied by TabPFN-Wide is omitted. Each model subsequently applies its own preprocessing, fitted on the training rows of each fold.

The four development datasets of HDLSS (madelon, arcene, RELATHE, and gisette, Section~\ref{sec:selection}) are excluded from the reported benchmark results, and their results are provided separately for completeness in Appendix~\ref{app:development}.

All models performed near chance on the LGG methylation task in MLOmics, despite prior evidence that DNA methylation distinguishes molecular glioma subtypes \citep{ceccarelli2016glioma}. We therefore audited the task for potential data inconsistencies, including sample–label correspondence, row order, duplicates, missing values, feature variability, and agreement between methylation and mRNA profiles from the same patients.

The identifiers, row order, labels, and feature values passed the corresponding checks, but the methylation profiles showed no concordance with the mRNA profiles of the same LGG patients. Across patient pairs, methylation-profile similarity correlated with mRNA-profile similarity in the other four cancers, with Spearman correlations ranging from 0.150 to 0.340, compared with absolute correlations of at most 0.004 after patient shuffling. In LGG, the correlation was $-0.012$, the same as with its patients shuffled. The mRNA, CNV, and miRNA profiles of LGG predicted the same labels with AUROCs ranging from 0.971 to 0.989 across all eight configurations in Table~\ref{tab:per_task_1}. These checks indicated an inconsistency involving the LGG methylation profiles but did not identify its cause. Since the problem could not be resolved from the released files, both the LGG methylation task and the combined LGG task, which contains the same methylation profiles, were excluded.

The reported benchmark results therefore cover 46 tasks: 11 from HDLSS, 23 from MLOmics, and 12 from Shamir. The 11 retained HDLSS datasets contain 2,000 to 22,283 features. These tasks are not independent samples, since different omics tasks from the same cancer share patients and labels, ALLAML and leukemia in HDLSS hold the same 72 patients, and some cohorts may overlap between benchmarks.

%% file: tables/benchmarks_description.tex
\begin{tabular}{llrrrrl} \toprule Benchmark & Dataset / Task group & Samples & Features & Classes & Min. class & CV \\ \midrule \begin{tabular}[t]{@{}l@{}} HDLSS \end{tabular} & ALLAML & 72 & 7,129 & 2 & 25 & $10\times3$ \\ & BASEHOCK & 1,993 & 4,862 & 2 & 994 & $10\times3$ \\ & CLL\_SUB\_111 & 111 & 11,340 & 3 & 11 & $10\times3$ \\ & GLI\_85 & 85 & 22,283 & 2 & 26 & $10\times3$ \\ & PCMAC & 1,943 & 3,289 & 2 & 961 & $10\times3$ \\ & Prostate\_GE & 102 & 5,966 & 2 & 50 & $10\times3$ \\ & SMK\_CAN\_187 & 187 & 19,993 & 2 & 90 & $10\times3$ \\ & TOX\_171 & 171 & 5,748 & 4 & 39 & $10\times3$ \\ & colon & 62 & 2,000 & 2 & 22 & $10\times3$ \\ & leukemia & 72 & 7,070 & 2 & 25 & $10\times3$ \\ & lung & 203 & 3,312 & 5 & 6 & $10\times3$ \\ \midrule \begin{tabular}[t]{@{}l@{}} MLOmics \end{tabular} & \begin{tabular}[t]{@{}l@{}} BRCA\\[-1pt] {\scriptsize mRNA, CNV, methylation, miRNA, combined} \end{tabular} & 671 & 345--57,133 & 5 & 31 & 5 \\ & \begin{tabular}[t]{@{}l@{}} COAD\\[-1pt] {\scriptsize mRNA, CNV, methylation, miRNA, combined} \end{tabular} & 260 & 375--56,204 & 4 & 4 & 5 \\ & \begin{tabular}[t]{@{}l@{}} GBM\\[-1pt] {\scriptsize mRNA, CNV, methylation, miRNA, combined} \end{tabular} & 244 & 308--56,417 & 5 & 20 & 5 \\ & \begin{tabular}[t]{@{}l@{}} LGG\\[-1pt] {\scriptsize mRNA, CNV, miRNA} \end{tabular} & 247 & 321--19,528 & 3 & 46 & 5 \\ & \begin{tabular}[t]{@{}l@{}} OV\\[-1pt] {\scriptsize mRNA, CNV, methylation, miRNA, combined} \end{tabular} & 284 & 244--56,147 & 4 & 64 & 5 \\ \midrule \begin{tabular}[t]{@{}l@{}} Shamir \end{tabular} & \begin{tabular}[t]{@{}l@{}} Breast\\[-1pt] {\scriptsize mRNA, methylation, miRNA, combined} \end{tabular} & 440 & 1,046--26,577 & 5 & 25 & 5 \\ & \begin{tabular}[t]{@{}l@{}} Glioblastoma\\[-1pt] {\scriptsize mRNA, methylation, miRNA, combined} \end{tabular} & 274 & 534--17,576 & 4 & 47 & 5 \\ & \begin{tabular}[t]{@{}l@{}} Sarcoma\\[-1pt] {\scriptsize mRNA, methylation, miRNA, combined} \end{tabular} & 254 & 1,046--26,577 & 7 & 6 & 5 \\ \bottomrule \end{tabular}

%% file: sections/05_training.tex
\section{Training}
\label{sec:training}

Pretraining starts from random initialization and proceeds in stages, with each subsequent stage continuing from the weights of the previous one. This section describes the pretraining objective and optimization, the stages of each model, how the models were selected, their size and pretraining budget compared with other TFMs, and the reproducibility of the released checkpoints.

\subsection{Objective and optimization}
\label{sec:optimization}

At each step, the model receives synthetic tables (Section~\ref{sec:prior}) and predicts the labels of their test rows from their labeled training rows in a single forward pass, as at inference. The objective is the cross-entropy between the predicted class probabilities and the true labels, averaged over the test rows.

Batches contain eight tables in the stages without widening, and a single table in the widening stages, where it is processed in one forward pass whether or not it is widened. In the widening stages, gradients are accumulated over eight batches per optimizer step so that the widest tables fit in memory. Since tables are generated continuously, an epoch is defined as 1,000 batches, which amounts to 8,000 tables and 1,000 optimizer steps in the stages without widening, and to 1,000 tables and 125 optimizer steps in the widening stages. Tables whose generation fails are discarded.

Optimization uses schedule-free AdamW \citep{defazio2024schedulefree}, without weight decay, with a linear warm-up of the learning rate over the first 500 optimizer steps of each stage and a new optimizer state at the start of each stage. Gradients are clipped to a norm of 1. Checkpoints store the averaged weights maintained by the optimizer, which are those used at inference. Training uses bfloat16 mixed precision and gradient checkpointing \citep{chen2016training}, which recomputes intermediate activations within each Transformer block during the backward pass instead of retaining them. Standardized feature values are clipped at $c=100$ (Section~\ref{sec:clip}).

\subsection{Training stages}
\label{sec:stages}

\begin{table}[t]
  \centering
  \caption{Pretraining stages of TabOmics v2 and v2.5. Feature and row ranges refer to tables sampled before feature widening. Rows are sampled log-uniformly. The widening probability is the fraction of batches augmented with 200–3,000 additional features, as described in Section~\ref{sec:widening}. Steps denote optimizer updates, and table counts are obtained from the training logs and rounded as displayed.
  }
  \label{tab:stages}
  \small
  \begin{tabular}{@{}llccccrrr@{}}
    \toprule
    Model & Stage & Features & Rows & Widened & Learning rate & Epochs & Steps & Tables \\
    \midrule
    v2   & 1. Base              & 2--100 & 40--1,024 & --  & $10^{-4}$        & 100 & 100k   & 0.80M \\
         & 2. Longer tables     & 2--100 & 40--2,048 & --  & $10^{-4}$        & 100 & 100k   & 0.80M \\
         & 3. Widening          & 2--350 & 40--1,024 & 0.3 & $10^{-4}$        & 30  & 3.75k  & 0.03M \\
         & 4. Low learning rate & 2--350 & 40--1,024 & 0.3 & $3\times10^{-5}$ & 30  & 3.75k  & 0.03M \\
         & Total                &        &           &     &                  &     & 207.5k & 1.66M \\
    \midrule
    v2.5 & 1. Base              & 2--100 & 40--1,024 & --  & $10^{-4}$        & 300 & 300k   & 2.40M \\
         & 2. Widening          & 2--350 & 40--1,024 & 0.3 & $10^{-4}$        & 30  & 3.75k  & 0.03M \\
         & 3. Low learning rate & 2--350 & 40--1,024 & 0.3 & $3\times10^{-5}$ & 30  & 3.75k  & 0.03M \\
         & Total                &        &           &     &                  &     & 307.5k & 2.46M \\
    \bottomrule
  \end{tabular}
\end{table}

Both models begin with a base stage on narrow tables of up to 100 features and 1,024 rows (Table~\ref{tab:stages}). The base stage of v2.5 continues the base run of v2 from epoch 100 to epoch 300, whereas v2 proceeds with 100 epochs on longer tables of up to 2,048 rows, a stage that v2.5 omits.

Both models then undergo two final stages with identical configurations and seeds, each starting from the preceding checkpoint of its own model. In the first, the prior generates tables with up to 350 features, and each batch is widened by 200 to 3,000 features with probability 0.3. The second keeps this data distribution and lowers the learning rate from $10^{-4}$ to $3\times10^{-5}$. Both stages are restricted to tables with at most 1,024 rows, so that widened tables of up to 3,350 features fit in GPU memory. Together, they add 7,500 optimizer steps and 60,000 tables to each model, such that most of the pretraining takes place in the narrow stages.

In total, v2 is trained for 207.5k optimizer steps on 1.66 million synthetic tables, and v2.5 for 307.5k steps on 2.46 million, with each stage running on a single GPU in the environment described in the \hyperref[sec:hardware_software]{Hardware and software environment} section. Each released model is the checkpoint at the end of its final stage, exported with the inference feature clip $c=10$. The configuration and seed of each stage are provided in the recipes of the \texttt{tabomics} package (Section~\ref{sec:reproducibility}).

\subsection{Model selection}
\label{sec:selection}

The training recipe and stage lengths of v2 were selected using a fixed validation protocol based on the loss on held-out synthetic tables, synthetic probes with known solutions, and development datasets. No reported benchmark task was included in this protocol. The development datasets comprise seven from the scikit-feature repository \citep{li2017featureselection} outside the benchmarks used here (GLIOMA, lymphoma, nci9, lung\_discrete, warpPIE10P, orlraws10P, and USPS) and three from the HDLSS benchmark (madelon, arcene, and RELATHE), set aside for development and excluded from all reported results.

The initial benchmark run with the selected v2 revealed a problem with extreme standardized values in very wide or sparse tables, including the benchmark datasets PCMAC and BASEHOCK. Diagnosis and development of the feature-clipping change were restricted to the development datasets listed above and gisette. This change lowered the feature clip at inference $c=100$, the value used in pretraining, to $c=10$ (Section~\ref{sec:inference}). The benchmarks were then run again, with gisette also set aside as a development dataset and excluded from the reported results.

TabOmics v2.5 was trained later, with a longer base stage and without the stage on longer tables. It was selected using the same fixed validation protocol, without using the reported benchmark results to choose the model. We release and report both models.

\subsection{Model size and pretraining budget}
\label{sec:budget}

\begin{table}[t]
  \centering
  \caption{Model size and reported pretraining budget of TabOmics and other TFMs. Tables is the reported number of synthetic tables processed during pretraining, with the TabOmics values taken from Table~\ref{tab:stages}. For TabOmics, compute includes only the stages leading to each released checkpoint. Compute values are reported as provided by the original studies and are not normalized across hardware, software environments, table dimensions, or numerical precision.
  }
  \label{tab:budget}
  \small
  \begin{tabular}{@{}lrrl@{}}
    \toprule
    Model & Parameters & Tables & Pretraining compute \\
    \midrule
    TabOmics v2  & 7.3M  & 1.66M & 2.1 A100 GPU-days \\
    TabOmics v2.5 & 7.3M & 2.46M & 2.4 A100 GPU-days \\
    TabPFNv2 \citep{hollmann2025tabpfn} & 7.2M\textsuperscript{a} & 130M & 2 weeks on 8 RTX 2080 Ti GPUs \\
    TabICLv2 \citep{qu2026tabiclv2} & 27.6M\textsuperscript{a} & 35M & 24.5 H100 GPU-days \\
    TabPFN-3.5 \citep{jaeger2026tabpfn35} & 220M & Not reported & Not reported \\
    \bottomrule
  \end{tabular}

  \smallskip
  \textsuperscript{a}\,Parameter counts from \citet{bilos2026mechanistic}.
\end{table}

TabOmics has approximately the same parameter count as TabPFNv2 and TabPFN-Wide, which continues pretraining from TabPFNv2, whereas TabICLv2 and TabPFN-3.5 have approximately 4 and 30 times as many parameters, respectively (Table~\ref{tab:budget}). TabPFNv2 was pretrained on approximately 50--80 times as many synthetic tables as the TabOmics models, with up to 2,048 training rows per table \citep{hollmann2025tabpfn}, and TabICLv2 on approximately 14--21 times as many, with 1,024 rows per table in its first stage and up to 60,000 in its last \citep{qu2026tabiclv2}.

Because table dimensions differ between models, the number of tables is only a rough measure of the pretraining budget. Reported GPU time provides additional context, although differences in hardware and implementation prevent a direct comparison of computational cost. The budgets of the two TabOmics models overlap, as both include the first 100 epochs of the shared base run. Section~\ref{sec:results_performance} relates the predictive performance of each model to its size.

\subsection{Reproducibility of the released checkpoints}
\label{sec:reproducibility}
 
The \texttt{tabomics} package includes the recipes of both models, with one configuration file per stage containing the recorded settings and seeds of the original runs. These recipes document the training procedure, but rerunning them does not reproduce the released weights bit for bit, for two reasons. First, the original GPU runs were not configured to ensure bitwise determinism. Second, all stages of v2 and the base stage of v2.5 were run before the Python hash seed was fixed, which prevents exact regeneration of their synthetic tables from the recorded random seeds alone.
 
The step counts, tables seen, and compute reported here are taken from the logs of the original runs. The released artifacts thus support procedural reproduction of the training, but not bitwise reconstruction of the published checkpoints. The evaluation, in contrast, can be reproduced from the released weights, which are the exact checkpoints evaluated in this paper: the experiment repository contains the fixed software environment, settings, logs, and per-fold outputs used to generate the tables and figures.

%% file: sections/06_interpretability.tex
\section{Interpretability and representations}
\label{sec:interpretability}

Besides predictions, a forward pass through TabOmics can return feature-importance scores and sample embeddings. The scores are derived from the attention of the label tokens, and the embeddings from their representations after the last block. This section describes both outputs, including the additional forward passes required to compute out-of-fold embeddings for the training rows.

\subsection{Feature importance}
\label{sec:importance}

Following TabPFN-Wide \citep{kolberg2025tabpfnwide, tabpfnwiderepo}, a feature score is derived from the attention paid by each label token to the corresponding feature token during feature-wise attention. These quantities are referred to as feature-importance scores. Let $a^{(l,h)}_{r,j}$ denote the attention weight from the label token of row $r$ to the token of feature $j$ in head $h$ of block $l$, with $H$ heads per block. The importance score of feature $j$ is
\begin{equation}
  s_j = \frac{1}{L\,H\,M} \sum_{l=1}^{L} \sum_{h=1}^{H} \sum_{r \in \mathcal{T}} a^{(l,h)}_{r,j},
  \label{eq:importance}
\end{equation}
where $\mathcal{T}$ is the set of the $M$ test rows. Attention from the label token to itself is excluded, and features dropped by preprocessing are left unscored. Since TabOmics assigns one token to each retained input feature, without grouping features or adding derived features, each score corresponds to a single input feature. TabPFN-Wide obtains this correspondence by disabling feature grouping and the other feature modifications of TabPFNv2 \citep{kolberg2025tabpfnwide}.

TabPFN-Wide averages these weights over both training and test rows, whereas TabOmics uses test rows only by default, because the label tokens serve different roles: those of training rows carry known labels, while those of test rows hold the representations decoded into predictions. Averaging over test rows therefore focuses the scores on the attention associated with the samples being predicted and avoids mixing training-row and test-row attention in proportions determined by their relative counts. In addition, for a fixed training context, the attention scores of a test row do not depend on the other test rows processed alongside it, so that the contribution of each test row to the scores is independent of the remaining ones. Averaging over all rows, as in TabPFN-Wide, or over training rows only remains available as an option.

The scores are extracted during the forward pass by explicitly computing only the attention weights from the label tokens, without materializing the full attention matrix (Section~\ref{sec:inference}). For fixed numbers of rows, blocks, and heads, this additional computation scales linearly with the number of features, whereas perturbation-based attribution methods generally require repeated model evaluations. When predictions are averaged over ensemble members, the scores are mapped back to the original feature order and averaged as well, which aims to reduce the sensitivity to column position. Since attention weights are not necessarily faithful explanations of a prediction \citep{jain2019attention, wiegreffe2019attention}, the scores are evaluated on synthetic tables with known informative features, and the predictive value, stability, and label concordance of the resulting gene rankings are assessed on cancer transcriptomic data (Sections~\ref{sec:importance_protocol} and~\ref{sec:results_importance}).

\subsection{Sample embeddings}
\label{sec:embeddings}

The embedding of a sample is the $E$-dimensional representation of its label token after the last block, immediately before the decoder. \citet{ye2025closerlook} study embeddings extracted from different layers of TabPFNv2: among the individual layers evaluated, the final layer achieves the best average rank, while combining layers selected on validation data further improves downstream classification. TabOmics uses only the final block, which avoids an additional layer-selection step.

For a test row, the embedding is available from the same forward pass as the prediction and is shaped by the labels of the training rows provided as context, while the label of the test sample remains hidden. A training row, however, carries its own label in its label token, so that its representation would depend on a label that is unavailable for a test row.

Following \citet{ye2025closerlook}, TabOmics therefore computes training embeddings out of fold. The training rows are split into class-stratified folds, 10 by default, and each fold is embedded as a set of test rows using the remaining folds as labeled context. Every training sample is thus embedded in the test-row role, with its own label hidden, although the available context differs across folds and from the full training context used to embed the actual test rows.

When multiple ensemble members are used to extract embeddings, they permute the columns and category codes but keep the class indices unchanged, which avoids changing the encoding of the context labels when combining representations across members. Test and out-of-fold training embeddings are both averaged across members, retaining dimension $E$. Sections~\ref{sec:embeddings_protocol} and~\ref{sec:results_embeddings} evaluate these representations as inputs to simple classifiers and for visualization.
 

%% file: sections/07_evaluation.tex
\section{Evaluation protocol}
\label{sec:evaluation}

This section describes the evaluated models, metrics, statistical tests, and ablations, together with the protocols used to assess feature-importance scores and sample embeddings. The code, settings, logs, and per-fold results of all experiments are available at \url{https://github.com/lbotecur/tabomics-paper}. 

\subsection{Models}
\label{sec:models}

TabOmics v2 and v2.5 use the default settings of the \texttt{tabomics} package: bfloat16 mixed precision, the feature clip $c=10$, and low-memory inference. They are compared with other TFMs and classical methods on the same folds, and each model uses its own preprocessing and the default settings of its package, except where stated:
\begin{itemize}
  \item \emph{TFMs}: TabPFN-3.5 \citep{jaeger2026tabpfn35}, TabPFNv2 \citep{hollmann2025tabpfn} with the checkpoint that encodes one feature per token (\texttt{gn2p4bpt}), TabPFN-Wide \citep{kolberg2025tabpfnwide} with its checkpoint trained with widening up to 5,000 features and one feature per group, and TabICLv2 \citep{qu2026tabiclv2}.
  \item \emph{Classical methods}: random forest \citep{breiman2001rf} and logistic regression with an elastic-net penalty (Elastic-net LR) \citep{zou2005elasticnet} using the defaults and implementations of scikit-learn\footnote{\url{https://github.com/scikit-learn/scikit-learn}} \citep{pedregosa2011scikit}, and XGBoost \citep{chen2016xgboost} with 1,000 trees and the histogram method as in TabPFN-Wide (using the XGBoost library\footnote{\url{https://github.com/dmlc/xgboost}}). For Elastic-net LR, a mixing parameter \texttt{l1\_ratio} of 0.5 is used, and its inverse regularization strength (\texttt{C}) is selected from five values by 3-fold cross-validation on the training rows, using log-loss as the criterion and fitting standardization within each partition. The solver converges in every fit within the limit of 5,000 iterations. As in TabPFN-Wide, random forest and XGBoost are run without hyperparameter search.
\end{itemize}

Every TFM, including TabOmics, is evaluated with one and eight ensemble members, denoted $\times 1$ and $\times 8$. Apart from the settings stated in this section, the number of members is the only setting changed from the defaults. The members of TabOmics are defined in Section~\ref{sec:inference}, whereas those of the other TFMs use the transformations implemented in their respective packages, which can also vary preprocessing and, for TabPFN-3.5, feature subsets.

TabPFN-3.5 processes up to 768 of the input features per member, distributed across members, so it sees up to about 6,000 with eight members. Of the 46 reported tasks, 40 have more than 768 features and 29 more than 6,000. For wide tables, the package recommends enough members to cover all features, but the same member count is retained for all TFMs so that they are compared at equal numbers of forward passes.

All tables are numeric, and no columns are explicitly declared categorical to any model. Automatic detection of categorical columns is disabled for TabOmics, whereas TabPFNv2 and TabPFN-3.5 retain their own rule, which treats columns with fewer than four distinct values as categorical when the table has at least 100 training rows. All models use the same random seed, pretrained weights are pinned by revision and SHA-256 hash, and package versions are fixed in the experiment repository. TFMs and XGBoost are evaluated on one GPU, whereas the other classical methods use the available CPU cores. The hardware configuration is specified in the \hyperref[sec:hardware_software]{Hardware and software environment} section.

\subsection{Metrics and statistics}
\label{sec:metrics}

AUROC and accuracy are reported. For binary tasks, AUROC is computed for class 1, and for multiclass tasks, one-vs-rest AUROC with a macro average over classes is used. When a class is absent from the test rows of a fold, AUROC is undefined and omitted, while accuracy is retained. This occurs in 5 of the 505 folds, one in each COAD task, identically for all models. Each metric is first averaged over the folds of a task and then over the tasks of a benchmark, giving equal weight to every task.

Each TabOmics configuration is compared with every other model using the per-task means and the Wilcoxon signed-rank test \citep{wilcoxon1945}, which assesses whether paired performance differences are centered at zero using their signs and the ranks of their absolute magnitudes. For each comparison, the numbers of tasks on which TabOmics performs better and worse are reported, together with the median difference. The $p$-values are adjusted using the Holm procedure \citep{holm1979}, which controls the family-wise error rate at 0.05. A family comprises the comparisons of one TabOmics configuration with the other models, separately for each metric and each set of tasks: all tasks or an individual benchmark.

Models are also ranked jointly within each task. The Friedman test assesses whether their ranks differ systematically, and the Nemenyi post hoc test determines the critical difference in average rank \citep{demsar2006}: two models differ significantly at the 0.05 level when their average ranks differ by more than this value. These tests use the per-task means of all 15 models, comprising four TabOmics configurations, eight configurations of the other TFMs, and three classical methods, on all 46 tasks. A non-significant difference does not establish equivalence and, since the tasks are not independent, the Holm correction controls multiplicity but not this dependence: the $p$-values are read as summaries of consistency across tasks, alongside the win counts and median differences, rather than as tests over independent cohorts.

Computational cost is measured as wall-clock time per fold, from model construction, including loading pretrained weights from local disk, to production of the test-row probabilities. This includes internal tuning, such as cross-validation of the elastic-net penalty. Timing is performed in a separate run with no competing workloads, and models are evaluated sequentially on the hardware specified in the \hyperref[sec:hardware_software]{Hardware and software environment} section, using the first fold of each reported task. Each measurement is repeated three times after one untimed run, which initializes the GPU where applicable and loads the weights for the first time. We report the median of the three runtimes and, for PyTorch-based GPU models, the peak GPU memory allocated through PyTorch. GPU memory is not reported for XGBoost or CPU-based methods because these allocations are not tracked by PyTorch.

\subsection{Ablations}
\label{sec:ablations}

Three ablations examine the inference choices described in Section~\ref{sec:inference}, with TabOmics v2.5 with one member as the reference configuration. Each ablation changes one setting and is evaluated on the same tasks and folds, using the same metrics and paired Wilcoxon tests with Holm correction across the ablation comparisons:
\begin{itemize}
  \item The feature clip of pretraining, $c=100$, instead of $c=10$.
  \item Single precision instead of bfloat16 mixed precision, assessed in terms of AUROC, accuracy, runtime, and memory.
  \item Two, four, or eight ensemble members instead of one, assessed in terms of AUROC, accuracy, runtime, and memory.
\end{itemize}

\subsection{Feature importance}
\label{sec:importance_protocol}

The importance scores of Equation~(\ref{eq:importance}) are assessed in three settings, with TabOmics v2.5 with one member (the reference configuration of Section~\ref{sec:ablations}) and TabPFN-Wide with the checkpoint and settings of Section~\ref{sec:models}. Since TabOmics adds a fixed random vector to each column position (Section~\ref{sec:encoding}), columns are randomly permuted in every repetition and fold, with the same permutation for all methods, to reduce the dependence of the rankings on fixed column positions.

\paragraph{Synthetic tables with known informative features.} Following the protocol of TabPFN-Wide, tables are generated with the \texttt{make\_classification} function of scikit-learn. Tables contain 200 to 5,000 features, of which 3, 5, or 7 are informative and the remainder are pure noise, with no redundant or repeated features, and five seeds are used per setting. This protocol is extended to 10,000, 20,000, and 30,000 features to cover widths comparable to the omics tasks.

Each table contains 250 samples, generated jointly and split into stratified sets of 50 training samples and 200 test samples. The training size follows TabPFN-Wide, while the test samples are added for this evaluation. TabOmics scores are extracted from the test samples, which also provide the predictive AUROC of each model, and the informative features are tracked through the column permutation.

Ranking quality is measured by feature recovery AUROC, the probability that an informative feature receives a higher score than a noise feature, with ties counted as one half. TabOmics is compared with TabPFN-Wide, random forest impurity importance, and a variant of TabOmics that averages attention over all rows, as TabPFN-Wide does. At each width, comparisons use the Wilcoxon test paired by table, separately for recovery AUROC and predictive AUROC, with Holm correction across methods.

\paragraph{Predictive value of the top-ranked genes.} TabPFN-Wide manually examined the 10 highest-attention genes in four omics datasets against the literature \citep{kolberg2025tabpfnwide}. The rankings are instead assessed quantitatively through prediction on held-out rows, stability across folds, and concordance with a known marker set.

Gene rankings are evaluated on the mRNA tasks of MLOmics and Shamir, which are also used for the embedding evaluation. Each method ranks genes using only the outer training rows of the folds defined in Section~\ref{sec:benchmarks}, which are split into stratified subsets of 80\% context rows and 20\% inner query rows. TabOmics extracts scores from the inner query rows, while TabPFN-Wide uses all rows of the same forward pass, following its default procedure. No method uses the outer test rows to rank genes.

For each ranking, logistic regression on standardized features is fitted on all outer training rows using the top $k$ genes and evaluated on the outer test rows, for $k = 10, 50, 100$, and 500, with the same classifier for every method. The methods comprise TabOmics, TabPFN-Wide, random forest importance, the analysis-of-variance (ANOVA) $F$ statistic, random gene selection with five draws, and an all-gene baseline. For each $k$, TabOmics is compared with every other method using the Wilcoxon test on per-task means, with Holm correction across methods.

Ranking stability across folds is measured by the Jaccard index of the top 100 genes and the Spearman correlation of the importance scores, both aligning genes by identifier.

\paragraph{Concordance with subtype markers.} The breast cancer labels in MLOmics and Shamir are PAM50 subtypes, assigned from the expression of the PAM50 genes \citep{parker2009pam50}. Recovering these genes therefore measures concordance with the definition of the labels, rather than providing independent biological validation. The PAM50 genes are counted among the top 10, 50, and 100 genes of each method and compared with chance expectations under a hypergeometric draw from the genes available in the task, with a one-sided test at the 0.05 level in each fold.

\subsection{Sample embeddings}
\label{sec:embeddings_protocol}

Embeddings are evaluated on the mRNA tasks of MLOmics and Shamir, which contain 12,042 to 20,531 genes and are also used for gene ranking. Since these benchmarks may share patients, they do not necessarily provide independent validations.

In each outer fold, test rows are embedded using all training rows as context. Training rows are embedded out of fold (Section~\ref{sec:embeddings}), using 10 stratified inner folds, each embedded with the other nine as context. This retains 90\% of the training rows as context for each inner fold, close to the full context used for test rows, and requires 10 out-of-fold forward passes per ensemble member and outer fold. The same inner folds are used for every model.

Each representation is evaluated with two simple classifiers chosen to probe its structure: logistic regression with an $\ell_2$ penalty, which assesses linear class separation, and a nearest-neighbor classifier with Euclidean distance and uniform weights, which assesses whether neighboring samples share their class. Every representation is standardized using training rows. Hyperparameters are selected on the outer training rows by stratified 3-fold cross-validation using macro AUROC, with the same search budget for every representation: five values of the inverse regularization strength, logarithmically spaced between $10^{-4}$ and $10^{4}$, and 1, 3, 5, or 11 neighbors are tested. The selected classifiers are refitted on all outer training rows and evaluated on the outer test rows.

For embeddings, this inner cross-validation uses the out-of-fold representations of the outer training rows, computed once per outer fold. Consequently, the labels of inner validation rows contribute to the context used to embed other training rows, which may make inner-validation scores optimistic and affect hyperparameter selection. No outer test label, however, contributes to a representation or to hyperparameter selection.

Embeddings from TabOmics v2.5 with one and eight members, the latter averaged across members without permuting the class indices (Section~\ref{sec:embeddings}), are compared with single-member embeddings from TabPFN-Wide and TabPFNv2, configured as in Section~\ref{sec:models}. All models use the same out-of-fold extraction procedure \citep{ye2025closerlook}, and for all of them the embedding is the representation of the label token after the last block, immediately before the decoder. TabPFN-Wide and TabPFNv2 are evaluated with one member because their package returns one embedding per member, each with its own permutation of the class indices, so averaging them would mix different encodings of the context labels.

Three representations without a TFM are included: standardized gene expression, principal components (PCA), and partial least squares (PLS) components. PCA retains $\min(E, n-1)$ components, where $E=192$ is the embedding dimension of TabOmics and $n$ is the number of rows used to fit PCA. PLS provides a supervised linear reduction of gene expression using the class labels, and 2, 5, 10, or 20 components are considered, with their number selected alongside the classifier hyperparameters. Scaling, PCA, and PLS are fitted within each inner training partition and applied unchanged to its validation partition.

For each classifier, TabOmics v2.5 with one member is compared with every other representation using the Wilcoxon test on per-task means, with Holm correction across representations. Direct TFM predictions on the same folds are reported as a reference, and two-dimensional projections of the embeddings from one task are shown for visualization.

%% file: sections/08_results.tex
\section{Results}
\label{sec:results}

The results follow the evaluation protocol of Section~\ref{sec:evaluation}. This section reports the predictive performance and computational cost of the models and the ablations of TabOmics v2.5, and then evaluates its feature-importance scores and sample embeddings.

\subsection{Predictive performance}
\label{sec:results_performance}

TabOmics v2.5 achieves the highest mean AUROC over the 46 tasks with eight ensemble members. Table~\ref{tab:benchmarks} reports the mean AUROC of each model per benchmark and across all tasks, its average rank, and the pairwise comparisons of TabOmics v2.5 with one and eight members against each non-TabOmics model. Accuracy results are provided in Table~\ref{tab:benchmarks_accuracy} in the appendix.

\begin{table}[t]
  \caption{Predictive performance across the three benchmarks. Values are mean AUROC by benchmark and across all 46 tasks, together with the average within-task AUROC rank. W–L gives the number of tasks on which the indicated TabOmics v2.5 reference configuration obtains a strictly higher or lower mean AUROC than the comparison model, ties are omitted. $p$-values are from paired Wilcoxon signed-rank tests on task-level means and are Holm-adjusted within each comparison family. Models are ordered by overall mean AUROC. Bold indicates the highest displayed value in each column, including ties after rounding. Dashes denote comparisons not reported.
  }
  \label{tab:benchmarks}
  \centering
  \small
  \setlength{\tabcolsep}{4pt}
  \input{tables/benchmarks_roc_auc}
\end{table}

With eight members, TabOmics v2.5 achieves a mean AUROC of 0.943 and the best average rank. It outperforms TabPFN-3.5 with eight members on 32 of the 46 tasks (median difference +0.0011, $p=0.034$), and each of the other non-TabOmics models on 38 to 46 tasks ($p<0.001$). This comparison fixes the number of forward passes rather than feature coverage: on 29 tasks, eight TabPFN-3.5 members do not cover all input features, whereas each TabOmics member processes every retained feature. Most wins against TabPFN-3.5 with eight members occur on the omics benchmarks, with 16 wins and 7 losses on MLOmics ($p=0.092$) and 10 wins and 2 losses on Shamir ($p=0.052$), compared with 6 wins and 5 losses on HDLSS ($p=0.898$).

With one member, TabOmics v2.5 significantly outperforms every non-TabOmics model except TabPFN-3.5 with eight members. The difference against that configuration is not significant (20 wins, 26 losses, median difference $-0.0006$, $p=0.410$), although TabOmics v2.5 uses a single forward pass and TabPFN-3.5 eight. In accuracy, the ordering is similar, but the difference between TabOmics v2.5 and TabPFN-3.5, both with eight members, is not significant (24 wins, 22 losses, $p=0.365$).

A joint rank analysis of the 15 configurations leads to the same general ordering (average ranks in Table~\ref{tab:benchmarks}). The Friedman test rejects the null hypothesis of equal performance ($p<0.001$), and the Nemenyi test at the 0.05 level gives a critical difference of 3.16 ranks, so two configurations differ significantly only if their average ranks differ by more than this value. TabOmics v2.5 with eight members has the best average rank (2.76) and differs significantly from every non-TabOmics model except TabPFN-3.5 with eight members (3.89). Its rank is also not significantly different from those of the other three TabOmics configurations (4.13, 4.60, and 5.35). The Nemenyi test accounts for all 105 pairwise comparisons, whereas the Wilcoxon tests focus on comparisons against each TabOmics configuration, with the correction families defined in Section~\ref{sec:metrics}.

The differences between the best models are small in absolute terms. The median absolute difference in AUROC between TabOmics v2.5 and TabPFN-3.5, both with eight members, is 0.0024, and across all 15 configurations the median task has an AUROC range of 0.057. These small differences partly reflect the benchmarks themselves: on the mRNA tasks, the best models reach a mean AUROC of 0.973, and several of these tasks have subtype labels defined from gene expression, such as the PAM50 subtypes of breast cancer, so that the features include the information used to assign the labels. Larger differences occur on omics modalities that do not define the labels. On CNV, for example, TabOmics v2.5 with one member reaches a mean AUROC of 0.877, compared with 0.794 for TabPFNv2 with one member and 0.731 for Elastic-net LR.

TabOmics v2.5 reaches this performance with 7.3M parameters, approximately the same number as TabPFNv2 and TabPFN-Wide (Table~\ref{tab:budget}). With one member, its mean AUROC exceeds theirs by 0.043 and 0.015, respectively. TabICLv2 has approximately four times as many parameters and TabPFN-3.5 approximately 30 times as many, yet TabOmics v2.5 with one member outperforms TabICLv2 with eight members and shows no significant AUROC difference from TabPFN-3.5 with eight.

TabOmics v2.5 outperforms TabOmics v2 with the same number of members: on 31 of the 46 tasks with one member (median difference +0.0012, $p=0.004$) and on 35 of the 46 tasks with eight members ($p<0.001$). The remaining evaluations therefore use TabOmics v2.5.

\subsection{Computational cost}
\label{sec:results_cost}

With one member, TabOmics v2.5 takes a median of 15.5 seconds per fold across all tasks, ranging from a benchmark median of 4.7 seconds on HDLSS to 20.3 seconds on MLOmics. Table~\ref{tab:cost} reports the time and peak GPU memory of each model alongside its mean AUROC, measured as described in Section~\ref{sec:metrics}.

On the widest task, which combines the four omics of breast cancer in MLOmics and contains 57,133 features and 671 rows, TabOmics v2.5 takes 217 seconds with one member. Its runtime is similar to those of TabPFN-Wide and TabPFNv2, which also process every feature in each member: with one member, these models take median times of 11.4 and 9.1 seconds, respectively, and 201 and 140 seconds on the widest task. TabPFN-3.5 and TabICLv2 are the fastest TFMs with one member, both at 2.5 seconds, and with eight members TabPFN-3.5 takes 3.1 seconds, with each member processing at most 768 features (Section~\ref{sec:models}). TabOmics v2.5 with one member shows no significant AUROC difference from TabPFN-3.5 with eight members (Section~\ref{sec:results_performance}), but takes approximately five times as long, although this comparison involves different feature coverage, since TabOmics attends to every feature in every forward pass. With eight members, both TabOmics models take approximately 65 seconds, close to TabPFN-Wide with eight members (67.0 seconds), and 1,506 seconds on the widest task.

\begin{table}[t]
  \caption{Predictive performance and computational cost. The table reports mean AUROC across 46 tasks, median runtime per fold within each benchmark and across tasks, maximum runtime, and median and maximum peak GPU memory. Runtime includes model construction, loading locally stored weights, and predicting the test rows, and is the median of three timed runs on the first fold of each task. GPU models were evaluated on one GPU, with ensemble members run sequentially, using the hardware specified in the \hyperref[sec:hardware_software]{Hardware and software environment} section. GPU memory is the peak allocation recorded by PyTorch and is not available for XGBoost or CPU-based methods. CPU--GPU runtime comparisons are specific to the reported hardware and are not hardware-normalized. Models are ordered as in Table~\ref{tab:benchmarks}.
  }
  \label{tab:cost}
  \centering
  \small
  \setlength{\tabcolsep}{4pt}
  \input{tables/cost}
\end{table}

Both TabOmics models use a median of 3.1\,GB of GPU memory and at most 27.7\,GB, on the widest task. These values are the same with one and eight members, since members run sequentially with low-memory inference (Section~\ref{sec:inference}). TabPFN-Wide uses a median of 7.9\,GB, whereas TabICLv2 uses a median of 14.8\,GB with one member and 48.9\,GB with eight, reaching a maximum of 70.7\,GB. Among the classical methods, Elastic-net LR, whose penalty is selected by cross-validation, takes a median of 58.4 seconds, longer than TabOmics v2.5 with one member.

\subsection{Ablations}
\label{sec:results_ablations}

Neither the feature clip nor the numerical precision significantly changes AUROC, and each changes its mean by approximately $10^{-4}$ (Table~\ref{tab:ablations}). The feature clip of pretraining, $c=100$, leaves AUROC unchanged on 10 tasks, increases it on 16, and decreases it on 20 relative to the default $c=10$ ($p=0.247$). The feature clip at inference therefore serves as a safeguard against extreme standardized values (Section~\ref{sec:clip}), with little effect on mean performance in these benchmarks. Single precision gives a higher AUROC than bfloat16 mixed precision on 24 tasks and a lower AUROC on 18 ($p=0.837$), but takes 3.6 times as long and approximately twice the GPU memory, and mixed precision is therefore retained as the default because it has similar AUROC and requires less time and memory.

\begin{table}[t]
  \caption{Ablations of TabOmics v2.5 relative to its default one-member configuration. W–T–L gives the numbers of tasks with higher, equal, or lower performance than the default. $p$-values are from paired Wilcoxon signed-rank tests and are Holm-adjusted across ablations. Runtime and memory are measured as in Table~\ref{tab:cost}. The feature-clipping ablation was not retimed because it does not change the computational procedure. Mean AUROC and accuracy values are reported to four decimal places to show the small differences between configurations.
  }
  \label{tab:ablations}
  \centering
  \small
  \setlength{\tabcolsep}{4pt}
  \input{tables/ablations}
\end{table}

Increasing the number of members improves AUROC consistently, although the gains are small. Two, four, and eight members outperform one member on 34, 34, and 37 of the 46 tasks, respectively ($p<0.001$ for each comparison). Mean AUROC rises from 0.9406 with one member to 0.9419, 0.9421, and 0.9426, so two members provide approximately two thirds of the gain obtained with eight. In accuracy, only the improvement with eight members is significant ($p=0.033$). Runtime increases by factors of 1.4, 2.4, and 4.2, respectively, while peak GPU memory remains unchanged because members run sequentially. Within TabOmics, one member provides the lowest-cost operating point, whereas eight members provide the highest mean AUROC at approximately four times the median runtime.

\subsection{Feature importance}
\label{sec:results_importance}

All importance scores of TabOmics v2.5 in this section are computed with one member (Section~\ref{sec:importance_protocol}).

\begin{figure}[t]
  \centering
  \includegraphics[width=\linewidth]{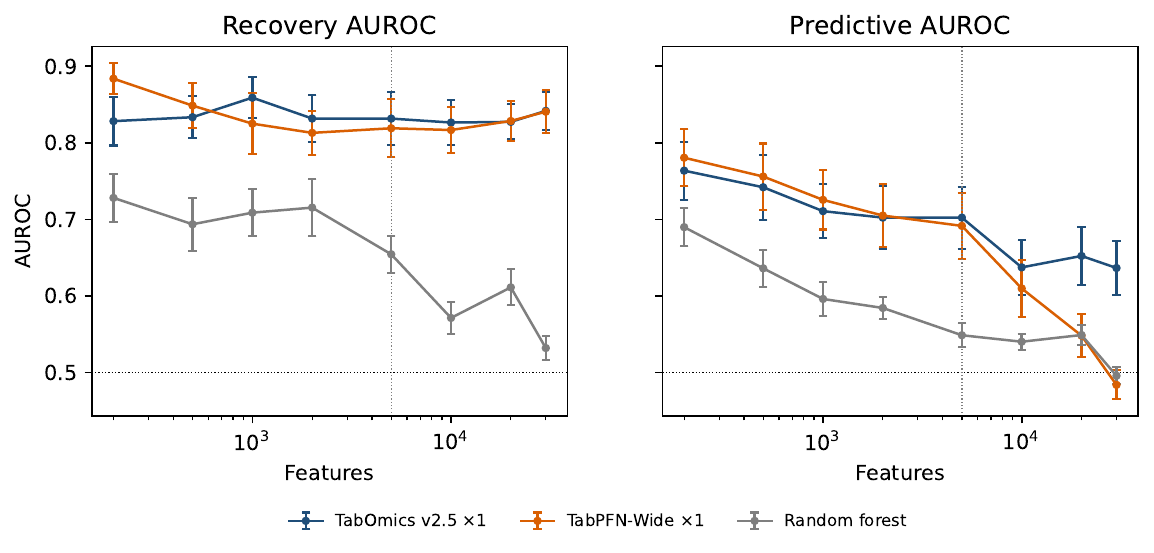}
  \caption{Feature recovery and predictive performance as table width increases. Left: AUROC for recovering the informative features from the feature-importance score rankings. Right: predictive AUROC on 200 held-out test rows. Points and error bars show the mean and standard error across 15 synthetic tables at each width, comprising five seeds for each setting with 3, 5, or 7 informative features. The horizontal dotted line indicates chance performance, and the vertical dotted line marks 5,000 features, the maximum width considered in the original TabPFN-Wide protocol.
  }
  \label{fig:importance_synthetic}
\end{figure}

\paragraph{Synthetic tables.} TabOmics v2.5 maintains a recovery AUROC between 0.826 and 0.859 as the number of features increases from 200 to 30,000 (Figure~\ref{fig:importance_synthetic}). Its recovery of informative features does not differ significantly from that of TabPFN-Wide at any width ($p \ge 0.083$). TabPFN-Wide has a higher mean recovery AUROC on the narrowest tables (0.884 versus 0.828 with 200 features), whereas from 2,000 features onward the means differ by at most 0.019. Random forest impurity importance recovers informative features less well, with a recovery AUROC that falls from 0.728 to 0.532, and TabOmics v2.5 outperforms it at every width, on 11 to 15 of the 15 tables ($p \le 0.045$).

Averaging attention over all rows, as in TabPFN-Wide, gives nearly the same recovery as averaging over test rows only. No width shows a significant difference, and the means differ by at most 0.002. In this experiment, the choice of rows therefore has little effect on recovery, although it changes which samples contribute to the scores.

The differences in predictive performance are larger than those in feature recovery. Up to 5,000 features, the two TFMs show no significant predictive AUROC difference at any width, with values decreasing from 0.764 to 0.702 for TabOmics v2.5 and from 0.781 to 0.691 for TabPFN-Wide. At widths beyond those used in its training, the AUROC of TabPFN-Wide falls to 0.548 with 20,000 features and to 0.484 with 30,000, whereas TabOmics v2.5 maintains an AUROC of 0.636 to 0.652 at these widths, outperforming TabPFN-Wide on 11 of the 15 tables with 20,000 features ($p=0.043$) and on 13 with 30,000 ($p=0.005$). Recovery and prediction therefore diverge for TabPFN-Wide: at 30,000 features, TabPFN-Wide retains a feature-recovery AUROC similar to that of TabOmics, whereas its mean predictive AUROC is 0.484.

\paragraph{Gene rankings.} On the eight mRNA tasks, logistic regression achieves similar mean AUROC when using the top-ranked genes of TabOmics v2.5, TabPFN-Wide, or random forest (Table~\ref{tab:importance_omics}). No value of $k$ shows a significant difference between TabOmics v2.5 and either of the other two methods ($p \ge 0.297$). Genes ranked by TabOmics v2.5 outperform random gene selection on all eight tasks for every $k$ ($p=0.031$, the smallest Holm-corrected $p$-value possible with eight tasks and four comparisons), and they also outperform selection by the ANOVA $F$ statistic on all eight tasks with 50 and 100 genes ($p=0.031$). With 50 genes, logistic regression using the TabOmics v2.5 ranking reaches a mean AUROC of 0.955, compared with 0.953 using all genes.

\begin{table}[t]
  \caption{Predictive value, cross-fold stability, and PAM50 concordance of gene rankings on the eight mRNA tasks. Predictive value is the mean AUROC of logistic regression fitted using the top $k$ genes from each outer-training ranking. The all-gene reference achieves 0.953. Stability is measured by the mean pairwise Jaccard index of the top 100 genes and the Spearman correlation between complete score vectors across folds. PAM50 concordance is the mean number of PAM50 genes among the top 10 in the two breast cancer tasks. The corresponding chance expectations are approximately 0.03 genes for MLOmics BRCA and 0.02 for Shamir breast. Bold indicates the highest displayed value among ranking methods. Dashes indicate that stability or marker concordance is not defined for random selection.}
  \label{tab:importance_omics}
  \centering
  \small
  \setlength{\tabcolsep}{4pt}
  \input{tables/importance_omics}
\end{table}

The rankings differ in stability across folds. The ANOVA $F$ statistic gives the most stable selection, with a Jaccard index of 0.61 for the top 100 genes, whereas TabOmics v2.5 and TabPFN-Wide have indices of 0.34 and 0.38, respectively, and random forest is the least stable at 0.14.

On the breast cancer tasks, whose labels are PAM50 subtypes, the top 10 genes ranked by TabOmics v2.5 contain a mean of 4.8 PAM50 genes on MLOmics and 4.2 on Shamir. Both are far above the counts expected by chance, about 0.03 and 0.02, respectively, with significant enrichment in every fold. TabPFN-Wide gives similar counts (4.2 and 4.4, respectively), whereas random forest finds fewer than half as many, and the ANOVA $F$ statistic finds a mean of 4.0 on MLOmics but 0.2 on Shamir. As discussed in Section~\ref{sec:importance_protocol}, these results measure concordance with the genes used to define the labels and therefore do not provide independent biological validation. The genes most often among the top 10 of TabOmics v2.5 in each breast cancer task are listed in Appendix~\ref{app:top_genes}.

\subsection{Sample embeddings}
\label{sec:results_embeddings}

TabOmics v2.5 is evaluated with one member unless eight are explicitly stated (Section~\ref{sec:embeddings_protocol}). With one or eight members, its embeddings achieve a higher mean AUROC than the representations of the other TFMs and those without a TFM, with both logistic regression and nearest neighbors (Table~\ref{tab:embeddings}). With logistic regression, the 192-dimensional embeddings reach a mean AUROC of 0.967, compared with 0.957 for the 12,042 to 20,531 input genes. The embeddings perform better on six of the eight tasks, although the difference is not significant ($p=0.445$).

With nearest neighbors, the improvement is larger. The embeddings reach a mean AUROC of 0.952, compared with 0.892 for the genes and 0.727 for the principal components, outperforming both on all eight tasks ($p=0.047$ for each comparison). The low AUROC of the principal components with nearest neighbors may partly reflect the standardization of every representation in the protocol, which gives each component the same scale in the distance calculation, including components that originally explain little variance. With either classifier, the embeddings of TabOmics v2.5 outperform those of TabPFN-Wide and TabPFNv2, and the partial least squares components, on six or seven of the eight tasks, although none of these differences is significant after correction for multiple comparisons.

\begin{table}[t]
  \caption{Classification using sample representations on the eight mRNA tasks. Values are mean AUROC for logistic regression and nearest-neighbor classifiers fitted to each representation. W–L gives the number of tasks on which TabOmics v2.5 ×1 performs better or worse than the comparison representation, and $p$-values are Holm-adjusted Wilcoxon test results. For TabOmics v2.5 ×8, embeddings are averaged across members. Direct TFM prediction AUROC on the same folds is included as a reference and is not applicable to gene-expression, PCA, or PLS representations. Bold indicates the highest displayed value in each AUROC column, including ties after rounding.}
  \label{tab:embeddings}
  \centering
  \small
  \setlength{\tabcolsep}{4pt}
  \input{tables/embeddings}
\end{table}

Classification through the embeddings retains much of the predictive performance of the TFMs. Logistic regression on the embeddings of TabOmics v2.5 reaches a mean AUROC of 0.967, compared with 0.972 for its direct predictions on the same folds, and for TabPFN-Wide both reach 0.963. For TabPFNv2, logistic regression on the embeddings gives a higher mean AUROC than its direct predictions (0.950 versus 0.937). Averaging the embeddings over eight members leaves the mean AUROC practically unchanged: 0.967 with logistic regression and 0.951 with nearest neighbors, against 0.967 and 0.952 with one member.

\begin{figure}[t]
  \centering
  \includegraphics[width=\linewidth]{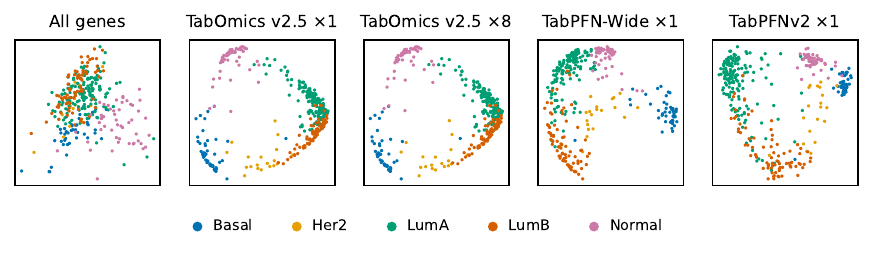}
  \caption{Two-dimensional PCA projections of gene-expression features and TFM sample embeddings. The panels show the 352 training samples from the first outer fold of the Shamir breast task, colored by PAM50 subtype. PCA was fitted separately to each standardized representation, and the first two principal components are shown. TFM training samples were represented using out-of-fold embeddings, so each sample was embedded without its own label in the context. The figure illustrates one fold. Quantitative results across all folds of the eight mRNA tasks are reported in Table~\ref{tab:embeddings}.
  }
  \label{fig:embeddings}
\end{figure}

Figure~\ref{fig:embeddings} illustrates the representations of the training rows in the first fold of Shamir breast, projected onto their first two principal components. In the standardized gene representation, the five PAM50 subtypes overlap in a single cloud, whereas in the embeddings of TabOmics v2.5, with either one or eight members, they follow an arc in a consistent order: Basal at one end, followed by Her2, LumB, LumA, and Normal at the other, with the two luminal subtypes adjacent. TabPFN-Wide and TabPFNv2 also separate Basal from the other subtypes. In TabPFN-Wide, Normal lies close to LumA, while in TabPFNv2, Her2 overlaps with LumB. This visualization illustrates one fold of one task, whereas the quantitative evaluation of class separation in the full representations is provided by the nearest-neighbor AUROC across all folds of the eight tasks (Table~\ref{tab:embeddings}).

%% file: tables/benchmarks_roc_auc.tex
\begin{tabular}{lrrrrrcrcr}
\toprule
 & \multicolumn{4}{c}{Mean AUROC} & & \multicolumn{2}{c}{TabOmics v2.5 $\times$1 vs.} & \multicolumn{2}{c}{TabOmics v2.5 $\times$8 vs.} \\
\cmidrule(lr){2-5} \cmidrule(lr){7-8} \cmidrule(lr){9-10}
Model & HDLSS & MLOmics & Shamir & All & Rank & W--L & $p$ & W--L & $p$ \\
\midrule
TabOmics v2.5 $\times$8 & 0.946 & \textbf{0.938} & \textbf{0.947} & \textbf{0.943} & \textbf{2.76} & -- & -- & -- & -- \\
TabPFN-3.5 $\times$8 & \textbf{0.948} & \textbf{0.938} & 0.945 & 0.942 & 3.89 & 20--26 & 0.410 & 32--14 & 0.034 \\
TabOmics v2 $\times$8 & 0.944 & 0.936 & \textbf{0.947} & 0.941 & 4.13 & -- & -- & -- & -- \\
TabOmics v2.5 $\times$1 & 0.945 & 0.935 & \textbf{0.947} & 0.941 & 4.60 & -- & -- & -- & -- \\
TabOmics v2 $\times$1 & 0.943 & 0.934 & 0.945 & 0.939 & 5.35 & -- & -- & -- & -- \\
TabPFN-Wide $\times$8 & 0.942 & 0.930 & 0.943 & 0.936 & 6.07 & 31--15 & $<$0.001 & 38--8 & $<$0.001 \\
TabPFN-3.5 $\times$1 & 0.929 & 0.926 & 0.933 & 0.929 & 8.72 & 37--9 & $<$0.001 & 40--6 & $<$0.001 \\
TabPFN-Wide $\times$1 & 0.940 & 0.915 & 0.933 & 0.926 & 9.59 & 39--7 & $<$0.001 & 43--3 & $<$0.001 \\
TabPFNv2 $\times$8 & 0.941 & 0.909 & 0.935 & 0.924 & 8.18 & 37--8 & $<$0.001 & 41--5 & $<$0.001 \\
TabICLv2 $\times$8 & 0.912 & 0.922 & 0.935 & 0.923 & 9.30 & 37--9 & $<$0.001 & 40--6 & $<$0.001 \\
TabICLv2 $\times$1 & 0.901 & 0.919 & 0.934 & 0.918 & 10.50 & 39--7 & $<$0.001 & 43--3 & $<$0.001 \\
XGBoost & 0.917 & 0.909 & 0.925 & 0.915 & 12.46 & 46--0 & $<$0.001 & 45--1 & $<$0.001 \\
Random forest & 0.927 & 0.905 & 0.920 & 0.914 & 12.11 & 45--1 & $<$0.001 & 45--1 & $<$0.001 \\
Elastic-net LR & 0.936 & 0.894 & 0.916 & 0.910 & 10.04 & 40--6 & $<$0.001 & 41--5 & $<$0.001 \\
TabPFNv2 $\times$1 & 0.925 & 0.876 & 0.915 & 0.898 & 12.30 & 44--2 & $<$0.001 & 46--0 & $<$0.001 \\
\bottomrule
\end{tabular}

%% file: tables/cost.tex
\begin{tabular}{lrrrrrrrr}
\toprule
 & & \multicolumn{5}{c}{Seconds per fold} & \multicolumn{2}{c}{GPU memory (GB)} \\
\cmidrule(lr){3-7} \cmidrule(lr){8-9}
Model & AUROC & HDLSS & MLOmics & Shamir & All & Max & Median & Max \\
\midrule
TabOmics v2.5 $\times$8 & 0.943 & 10.9 & 87.3 & 31.8 & 65.2 & 1,506 & 3.1 & 27.7 \\
TabPFN-3.5 $\times$8 & 0.942 & 2.4 & 3.2 & 3.2 & 3.1 & 16 & 1.6 & 6.4 \\
TabOmics v2 $\times$8 & 0.941 & 11.0 & 87.5 & 32.0 & 65.3 & 1,506 & 3.1 & 27.7 \\
TabOmics v2.5 $\times$1 & 0.941 & 4.7 & 20.3 & 8.1 & 15.5 & 217 & 3.1 & 27.7 \\
TabOmics v2 $\times$1 & 0.939 & 4.7 & 20.5 & 8.1 & 15.6 & 217 & 3.1 & 27.7 \\
TabPFN-Wide $\times$8 & 0.936 & 10.6 & 85.7 & 34.2 & 67.0 & 1,508 & 7.9 & 28.4 \\
TabPFN-3.5 $\times$1 & 0.929 & 2.1 & 2.6 & 2.2 & 2.5 & 11 & 1.6 & 6.4 \\
TabPFN-Wide $\times$1 & 0.926 & 2.5 & 13.9 & 6.2 & 11.4 & 201 & 7.9 & 28.4 \\
TabPFNv2 $\times$8 & 0.924 & 8.4 & 63.4 & 24.4 & 47.6 & 1,031 & 5.4 & 19.2 \\
TabICLv2 $\times$8 & 0.923 & 38.7 & 121.3 & 52.8 & 100.9 & 687 & 48.9 & 70.7 \\
TabICLv2 $\times$1 & 0.918 & 0.8 & 3.3 & 1.5 & 2.5 & 47 & 14.8 & 59.4 \\
XGBoost & 0.915 & 0.8 & 7.8 & 5.3 & 5.0 & 50 & -- & -- \\
Random forest & 0.914 & 2.2 & 6.6 & 3.1 & 5.5 & 21 & -- & -- \\
Elastic-net LR & 0.910 & 13.8 & 115.9 & 78.1 & 58.4 & 1,306 & -- & -- \\
TabPFNv2 $\times$1 & 0.898 & 2.3 & 10.5 & 4.3 & 9.1 & 140 & 5.4 & 19.2 \\
\bottomrule
\end{tabular}

%% file: tables/ablations.tex
\begin{tabular}{lrcrrcrrr}
\toprule
 & \multicolumn{3}{c}{AUROC} & \multicolumn{3}{c}{Accuracy} & & \\
\cmidrule(lr){2-4} \cmidrule(lr){5-7}
TabOmics v2.5 & Mean & W--T--L & $p$ & Mean & W--T--L & $p$ & Seconds & GPU (GB) \\
\midrule
One member (reference) & 0.9406 & -- & -- & 0.8277 & -- & -- & 15.5 & 3.1 \\
Feature clip 100 & 0.9405 & 16--10--20 & 0.247 & 0.8276 & 7--29--10 & 1.000 & -- & -- \\
Single precision & 0.9407 & 24--4--18 & 0.837 & 0.8279 & 10--25--11 & 1.000 & 56.0 & 6.0 \\
Two members & 0.9419 & 34--1--11 & $<$0.001 & 0.8294 & 27--5--14 & 0.341 & 21.3 & 3.1 \\
Four members & 0.9421 & 34--1--11 & $<$0.001 & 0.8299 & 28--2--16 & 0.341 & 36.7 & 3.1 \\
Eight members & 0.9426 & 37--0--9 & $<$0.001 & 0.8315 & 30--2--14 & 0.033 & 65.2 & 3.1 \\
\bottomrule
\end{tabular}

%% file: tables/importance_omics.tex
\begin{tabular}{lrrrrrrrr}
\toprule
 & \multicolumn{4}{c}{AUROC of the top $k$ genes} & \multicolumn{2}{c}{Stability} & \multicolumn{2}{c}{PAM50 in top 10} \\
\cmidrule(lr){2-5} \cmidrule(lr){6-7} \cmidrule(lr){8-9}
Method & $k=10$ & $k=50$ & $k=100$ & $k=500$ & Jaccard & Spearman & MLOmics & Shamir \\
\midrule
TabOmics v2.5 $\times$1 & 0.926 & 0.955 & 0.954 & 0.962 & 0.34 & 0.75 & \textbf{4.8} & 4.2 \\
TabPFN-Wide $\times$1 & \textbf{0.929} & \textbf{0.956} & \textbf{0.957} & \textbf{0.965} & 0.38 & 0.77 & 4.2 & \textbf{4.4} \\
Random forest & 0.920 & 0.949 & 0.953 & 0.959 & 0.14 & 0.08 & 2.0 & 1.6 \\
ANOVA F & 0.897 & 0.927 & 0.939 & 0.961 & \textbf{0.61} & \textbf{0.92} & 4.0 & 0.2 \\
Random genes & 0.782 & 0.884 & 0.911 & 0.945 & -- & -- & -- & -- \\
\bottomrule
\end{tabular}

%% file: tables/embeddings.tex
\begin{tabular}{lrcrrcrr}
\toprule
 & \multicolumn{3}{c}{Logistic regression} & \multicolumn{3}{c}{Nearest neighbors} & Direct \\
\cmidrule(lr){2-4} \cmidrule(lr){5-7} \cmidrule(lr){8-8}
Representation & AUROC & W--L & $p$ & AUROC & W--L & $p$ & AUROC \\
\midrule
TabOmics v2.5 $\times$1 & \textbf{0.967} & -- & -- & \textbf{0.952} & -- & -- & 0.972 \\
TabOmics v2.5 $\times$8 & \textbf{0.967} & 3--5 & 0.742 & 0.951 & 5--2 & 0.297 & \textbf{0.973} \\
TabPFN-Wide $\times$1 & 0.963 & 7--1 & 0.445 & 0.948 & 6--2 & 0.297 & 0.963 \\
TabPFNv2 $\times$1 & 0.950 & 7--1 & 0.156 & 0.932 & 7--1 & 0.164 & 0.937 \\
All genes & 0.957 & 6--2 & 0.445 & 0.892 & 8--0 & 0.047 & -- \\
PLS & 0.945 & 7--1 & 0.141 & 0.927 & 7--1 & 0.094 & -- \\
PCA & 0.939 & 7--1 & 0.141 & 0.727 & 8--0 & 0.047 & -- \\
\bottomrule
\end{tabular}

%% file: sections/09_limitations.tex
\section{Discussion and limitations}
\label{sec:limitations}

\paragraph{Interpretation.} With one member, TabOmics v2.5 reaches a mean AUROC of 0.941, comparable to the 0.942 of TabPFN-3.5 with eight members ($p=0.410$, Table~\ref{tab:benchmarks}). This may reflect the fact that each TabOmics forward pass processes all retained features jointly, whereas each TabPFN-3.5 member receives only a subset of the features when their number exceeds its per-member capacity. The explanation remains conjectural, as no ablation isolates the contribution of joint feature processing. The comparison also carries a computational cost: the median runtime per fold is approximately five times higher for TabOmics v2.5 with one member (15.5 seconds) than for TabPFN-3.5 with eight (3.1 seconds, Section~\ref{sec:results_cost}).

\paragraph{Additional outputs.} The same pretrained model also returns feature-importance scores and sample embeddings, without task-specific parameter updates. On synthetic tables, the scores rank informative variables above noise variables, and on the eight mRNA tasks, logistic regression on the embeddings reaches a mean AUROC of 0.967, compared with 0.972 for direct prediction (Sections~\ref{sec:results_importance} and~\ref{sec:results_embeddings}). 

\paragraph{Model.} TabOmics processes every retained feature jointly in each forward pass. This allows the model to use all genes after preprocessing, but makes it slower than models that subsample features. Because attention weights are not materialized, the memory required for attention between features grows linearly with the number of features, whereas computation remains quadratic. On the widest task, with 57,133 features, the model takes 217 seconds with one member and 1,506 seconds with eight (Section~\ref{sec:results_cost}). The released models support classification with up to 10 classes. The architecture also supports regression and fine-tuning, but no regression model is released and fine-tuning is not evaluated here. The released recipes cannot reproduce the released checkpoints exactly, although code, weights, and recipes are publicly available (Section~\ref{sec:reproducibility}).

\paragraph{Attribution of gains.} The evaluated variations (clipping, precision, and ensemble size) do not isolate the contribution of the TabICLv2 prior or of the feature-widening procedure. Whether the observed performance is due to either component, or to their combination, therefore remains untested. A control trained without widening under an equal compute budget would be required to answer this question, and it is left for future work.

\paragraph{Benchmark design.} Several tasks are close to their apparent performance ceiling, particularly the mRNA tasks, whose subtype labels are derived from gene-expression measurements. This limits the differences that can be observed among the strongest models (Section~\ref{sec:results_performance}). The 46 tasks should also not be interpreted as 46 independent cohorts: different omics modalities of the same cancer type share patients and labels, and cohorts may overlap across benchmarks. The statistical tests therefore summarize the consistency of the observed differences across tasks, not across independent patient cohorts.

\paragraph{Baselines.} Random forest and XGBoost were run without hyperparameter search, following the TabPFN-Wide protocol. Although TabPFN-3.5 recommends using enough ensemble members to cover all features in wide tables—which would require over 50 forward passes on our widest datasets—all TFMs were evaluated under the same member count to ensure a controlled comparison with an equal budget of forward passes. Consequently, with eight members, TabPFN-3.5 does not cover all retained features in 29 of the 46 tasks (Section~\ref{sec:models}), whereas TabOmics evaluates every retained feature jointly within each forward pass. The comparison thus characterizes this fixed-budget operating point and does not indicate the maximum performance TabPFN-3.5 could attain with full feature coverage.

\paragraph{Scope.} TabOmics was designed and evaluated for high-dimensional, low-sample-size tables, and its performance on narrow tabular datasets is unknown. The tasks considered here predict tumor subtypes or histological types, not patient outcomes. The results therefore do not establish clinical utility for outcomes such as survival, disease progression, or treatment response, which would require evaluation on independent cohorts and clinically relevant endpoints.

\paragraph{Feature-importance scores and biological interpretation.} Attention weights are not necessarily faithful explanations of model predictions \citep{jain2019attention, wiegreffe2019attention}. The outputs of TabOmics should therefore be read as feature-importance scores, not as causal or mechanistic feature importance. On synthetic tables, the feature-recovery AUROCs range from 0.826 to 0.859. This supports their empirical utility under known ground truth but does not show that the scores explain individual predictions faithfully. In the breast cancer tasks, the recovery of PAM50 genes indicates concordance with the molecular information used to define the subtype labels. It is a positive control, not independent biological validation. None of the highly ranked genes identified in this study was experimentally validated, and their biological or clinical relevance cannot be inferred from the rankings alone. Gene rankings and sample embeddings were evaluated on only eight mRNA tasks, which limits the statistical power of the paired Wilcoxon tests to distinguish among methods (Sections~\ref{sec:results_importance} and~\ref{sec:results_embeddings}).

\paragraph{Sample embeddings.} Training-sample representations were computed out of fold, so each training sample was embedded without its own label in the context. The embeddings used for inner hyperparameter selection, however, were computed once per outer fold. The labels of inner-validation samples may therefore contribute to the context used to embed inner-training samples, which may make the selection of downstream-classifier hyperparameters optimistic. The labels of the outer test samples are never used to construct their representations or to select those hyperparameters (Section~\ref{sec:embeddings_protocol}).

%% file: sections/10_conclusion.tex
\section{Conclusion}
\label{sec:conclusion}

TabOmics is a 7.3M-parameter TFM pretrained exclusively on synthetic tables with a budget of a few GPU-days. It processes all features retained from wide omics tables jointly and achieves competitive classification performance against models with up to 30 times as many parameters. Across 46 tasks from three benchmarks of high-dimensional, low-sample-size tables, TabOmics v2.5 with eight ensemble members obtains the highest mean AUROC among the evaluated configurations (0.943). With one member, its mean AUROC (0.941) is comparable to that of TabPFN-3.5 with eight members (0.942), at the cost of a longer inference time.

The model also returns attention-derived feature-importance scores and sample embeddings, with test-sample embeddings obtained from the same forward pass as the predictions. On synthetic tables, the scores rank informative features above noise with a recovery AUROC similar to that of TabPFN-Wide, including at widths beyond those seen during pretraining. 
Under this evaluation protocol, logistic regression on the embeddings achieves a mean AUROC of 0.967, compared with a mean AUROC of 0.972 for direct prediction. Nearest-neighbor classification on the embeddings achieves a mean AUROC of 0.952, compared with 0.892 for nearest-neighbor classification on gene-expression features.

The source code, model weights, synthetic-data generator, staged training recipes, evaluation scripts, and per-fold results are publicly available. They allow the study of synthetic pretraining priors, the computational cost of joint feature attention, and the information encoded in feature-importance scores and sample embeddings. Future work includes ablations that isolate the effect of joint feature attention and of the feature-widening procedure under an equal compute budget, evaluation on additional omics modalities, and evaluation on clinically relevant outcomes such as survival and treatment response.

\section*{Hardware and software environment}
\phantomsection 
\label{sec:hardware_software}

The experiments were run on a single machine with eight NVIDIA A100-SXM4 GPUs, each with 80\,GB of memory, and two AMD EPYC 7282 16-core CPUs, providing 32 CPU cores in total. Each experimental run used at most one GPU. The machine ran Linux 4.18.0-553.148.1.el8\_10.x86\_64 with NVIDIA driver 550.163.01. The experiments used Python 3.12.15, PyTorch 2.9.0 with CUDA 12.8 support (\texttt{torch 2.9.0+cu128}), CUDA 12.8, and cuDNN 9.10.2. 

The software environment used to generate the reported results is fixed in the experiment repository. All TabOmics experiments use version 0.3.0 of the \texttt{tabomics} package.

\section*{Availability and licenses} 

The \texttt{tabomics} package is free software released under the GNU General Public License version 3.0 or later (GPL-3.0-or-later). Its source code, documentation, synthetic-data generator, and staged training recipes for TabOmics v2 and v2.5 are available at \url{https://github.com/lbotecur/tabomics}. All experimental artifacts associated with this paper are available at \url{https://github.com/lbotecur/tabomics-paper} and are released under the same license as the \texttt{tabomics} package. The evaluated weights of TabOmics v2 and v2.5 are available at \url{https://huggingface.co/lbotecur/tabomics-v2} and \url{https://huggingface.co/lbotecur/tabomics-v2.5}, respectively. 

The TabOmics architecture builds on nanoTabPFN \citep{pfefferle2025nanotabpfn,nanotabpfnrepo} and its implementation in TFM-Playground \citep{tfmplayground}, both released under the Apache License 2.0. The synthetic-table generator adapts the prior implemented in the \texttt{tabicl} package \citep{qu2025tabicl, qu2026tabiclv2,tabicl2026software}, released under the BSD 3-Clause License. The fine-tuning implementation also follows the \texttt{tabicl} package and \citet{rubachev2025finetuning}. The feature-widening procedure and attention-derived feature-importance calculation follow the methods described for TabPFN-Wide \citep{kolberg2025tabpfnwide,tabpfnwiderepo}, and the out-of-fold sample-embedding procedure follows \citet{ye2025closerlook}. These methods were reimplemented in \texttt{tabomics}. No TabPFN-Wide model weights are redistributed with the package.
 
\section*{Acknowledgments}

This work was supported by the Spanish Ministry of Science, Innovation and Universities

(MICIU/AEI/10.13039/501100011033) through project grant AIA2025-163540-C32.

%% file: sections/appendix.tex
\section{Per-task results}
\label{app:per_task}

Table~\ref{tab:benchmarks_accuracy} reports the mean accuracy of each model per benchmark and over the 46 tasks, as Table~\ref{tab:benchmarks} does for AUROC. Tables~\ref{tab:per_task_1} and~\ref{tab:per_task_2} report the AUROC among all 15 models for each task, averaged over its folds. In each COAD task, one fold lacks a class in its test rows and is excluded from the AUROC (Section~\ref{sec:metrics}).

\begin{table}[h]
  \caption{Mean accuracy per benchmark and over the 46 tasks, average rank, and pairwise comparisons, as in Table~\ref{tab:benchmarks} for AUROC.}
  \label{tab:benchmarks_accuracy}
  \centering
  \small
  \setlength{\tabcolsep}{4pt}
  \input{tables/benchmarks_accuracy}
\end{table}

\begin{table}[p]
  \caption{AUROC of both TabOmics models, TabPFN-3.5, and TabPFN-Wide, with one and eight members, for each task, averaged across folds. Bold indicates the highest displayed AUROC among all 15 models for each task, including ties after rounding. The highest value may appear in this table, in Table~\ref{tab:per_task_2}, or in both.}
  \label{tab:per_task_1}
  \centering
  \footnotesize
  \setlength{\tabcolsep}{4pt}
  \input{tables/per_task_1}
\end{table}

\begin{table}[p]
  \caption{AUROC of TabPFNv2, TabICLv2, and the classical methods for each task, averaged across folds. Bold indicates the highest displayed AUROC among all 15 models for each task, including ties after rounding. The highest value may appear in this table, in Table~\ref{tab:per_task_1}, or in both.}
  \label{tab:per_task_2}
  \centering
  \footnotesize
  \setlength{\tabcolsep}{4pt}
  \input{tables/per_task_2}
\end{table}

\section{Development datasets}
\label{app:development}

The four development datasets of the HDLSS benchmark are excluded from the reported results (Section~\ref{sec:selection}). Of these, madelon, arcene, and RELATHE were used to select the training recipes, and all four were included in the subsequent development checks. Table~\ref{tab:development} reports the AUROC of the 15 models on these datasets, on the folds of the benchmark. TabOmics v2.5 with one member reaches a mean AUROC of 0.947, below every TFM outside TabOmics except TabICLv2. All four TabOmics configurations are weakest on madelon, a synthetic dataset whose label depends on interactions between a few features, with AUROCs of 0.886 to 0.898, compared with 0.939 to 0.970 for the other TFMs.

Against TabPFN-3.5, TabPFN-Wide, and TabPFNv2, the median AUROC difference between TabOmics v2.5 with one member and each model is lower on the development datasets than on the reported tasks, and it is negative against TabPFN-3.5 and TabPFN-Wide. Against TabICLv2 and the classical methods, it is higher on the development datasets than on the reported tasks, by 0.0010 to 0.0208. On the datasets used for its selection, TabOmics v2.5 therefore does not outperform the strongest models.

\begin{table}[h]
  \caption{AUROC on the four development datasets, averaged over the folds of the HDLSS benchmark, and median AUROC difference between TabOmics v2.5 with one member and each model on these datasets (Development) and on the 46 reported tasks (Reported). Models are sorted by their mean AUROC over the four datasets. The highest value in each AUROC column is in bold.}
  \label{tab:development}
  \centering
  \small
  \setlength{\tabcolsep}{4pt}
  \input{tables/development}
\end{table}

\section{Top-ranked genes of the breast cancer tasks}
\label{app:top_genes}

Table~\ref{tab:top_genes} lists, for the two breast cancer tasks of Section~\ref{sec:results_importance}, the 10 genes that appear most often among the top 10 of TabOmics v2.5 with one member across the five folds. Six of these 10 genes are PAM50 genes in each task, whereas about 0.03 on MLOmics and 0.02 on Shamir would be expected by chance among 10 genes. KRT5 and ESR1 appear in both lists.

\begin{table}[h]
  \caption{Genes most frequently ranked among the top 10 by TabOmics v2.5 ×1 in the breast cancer tasks. For each task, the table reports the 10 genes with the highest selection frequency across the five outer folds. Ties are resolved by lower mean rank. PAM50 genes are shown in bold.}
  \label{tab:top_genes}
  \centering
  \small
  \input{tables/top_genes}
\end{table}

%% file: tables/benchmarks_accuracy.tex
\begin{tabular}{lrrrrrcrcr}
\toprule
 & \multicolumn{4}{c}{Mean accuracy} & & \multicolumn{2}{c}{TabOmics v2.5 $\times$1 vs.} & \multicolumn{2}{c}{TabOmics v2.5 $\times$8 vs.} \\
\cmidrule(lr){2-5} \cmidrule(lr){7-8} \cmidrule(lr){9-10}
Model & HDLSS & MLOmics & Shamir & All & Rank & W--L & $p$ & W--L & $p$ \\
\midrule
TabOmics v2.5 $\times$8 & 0.882 & \textbf{0.827} & \textbf{0.795} & \textbf{0.832} & \textbf{3.53} & -- & -- & -- & -- \\
TabPFN-3.5 $\times$8 & \textbf{0.896} & 0.820 & 0.790 & 0.830 & 3.62 & 19--27 & 0.585 & 24--22 & 0.365 \\
TabOmics v2 $\times$8 & 0.880 & 0.822 & 0.794 & 0.829 & 4.13 & -- & -- & -- & -- \\
TabOmics v2.5 $\times$1 & 0.881 & 0.819 & \textbf{0.795} & 0.828 & 4.75 & -- & -- & -- & -- \\
TabOmics v2 $\times$1 & 0.877 & 0.818 & 0.792 & 0.825 & 5.71 & -- & -- & -- & -- \\
TabPFN-Wide $\times$8 & 0.887 & 0.803 & 0.790 & 0.820 & 6.13 & 30--15 & 0.004 & 31--11 & $<$0.001 \\
TabPFN-3.5 $\times$1 & 0.860 & 0.803 & 0.760 & 0.806 & 8.37 & 36--10 & $<$0.001 & 39--6 & $<$0.001 \\
TabPFN-Wide $\times$1 & 0.879 & 0.775 & 0.767 & 0.798 & 9.40 & 38--7 & $<$0.001 & 39--4 & $<$0.001 \\
TabPFNv2 $\times$8 & 0.885 & 0.772 & 0.763 & 0.797 & 8.22 & 37--8 & $<$0.001 & 38--6 & $<$0.001 \\
TabICLv2 $\times$8 & 0.814 & 0.787 & 0.757 & 0.786 & 9.88 & 40--6 & $<$0.001 & 42--2 & $<$0.001 \\
XGBoost & 0.841 & 0.777 & 0.749 & 0.785 & 11.51 & 44--2 & $<$0.001 & 44--2 & $<$0.001 \\
Elastic-net LR & 0.876 & 0.760 & 0.733 & 0.781 & 9.98 & 38--8 & $<$0.001 & 36--8 & $<$0.001 \\
Random forest & 0.857 & 0.770 & 0.726 & 0.779 & 11.73 & 42--3 & $<$0.001 & 45--1 & $<$0.001 \\
TabICLv2 $\times$1 & 0.788 & 0.781 & 0.758 & 0.777 & 10.73 & 43--3 & $<$0.001 & 43--3 & $<$0.001 \\
TabPFNv2 $\times$1 & 0.842 & 0.639 & 0.727 & 0.710 & 12.32 & 40--6 & $<$0.001 & 43--2 & $<$0.001 \\
\bottomrule
\end{tabular}

%% file: tables/per_task_1.tex
\begin{tabular}{lrrrrrrrr}
\toprule
Task & \multicolumn{2}{c}{TabOmics v2.5} & \multicolumn{2}{c}{TabOmics v2} & \multicolumn{2}{c}{TabPFN-3.5} & \multicolumn{2}{c}{TabPFN-Wide} \\
\cmidrule(lr){2-3} \cmidrule(lr){4-5} \cmidrule(lr){6-7} \cmidrule(lr){8-9}
 & $\times$1 & $\times$8 & $\times$1 & $\times$8 & $\times$1 & $\times$8 & $\times$1 & $\times$8 \\
\midrule \multicolumn{9}{l}{\emph{HDLSS}} \\
ALLAML & \textbf{0.998} & \textbf{0.998} & \textbf{0.998} & \textbf{0.998} & 0.989 & 0.996 & 0.996 & 0.997 \\
BASEHOCK & 0.998 & \textbf{0.999} & 0.996 & 0.998 & 0.974 & \textbf{0.999} & \textbf{0.999} & \textbf{0.999} \\
CLL\_SUB\_111 & 0.897 & \textbf{0.898} & 0.893 & 0.896 & 0.821 & 0.891 & 0.875 & 0.883 \\
GLI\_85 & 0.944 & 0.945 & 0.939 & 0.945 & 0.944 & \textbf{0.951} & 0.945 & 0.947 \\
PCMAC & 0.975 & 0.983 & 0.975 & 0.978 & 0.947 & \textbf{0.987} & 0.979 & 0.979 \\
Prostate\_GE & 0.969 & \textbf{0.970} & 0.967 & 0.967 & 0.964 & 0.969 & 0.962 & 0.959 \\
SMK\_CAN\_187 & 0.765 & 0.764 & \textbf{0.769} & 0.766 & 0.739 & 0.758 & 0.767 & 0.765 \\
TOX\_171 & 0.982 & 0.982 & 0.978 & 0.980 & 0.971 & \textbf{0.988} & 0.963 & 0.972 \\
colon & 0.874 & 0.873 & 0.866 & 0.864 & 0.883 & \textbf{0.889} & 0.864 & 0.871 \\
leukemia & 0.998 & 0.998 & \textbf{0.999} & \textbf{0.999} & 0.987 & 0.998 & 0.997 & 0.997 \\
lung & \textbf{0.997} & \textbf{0.997} & \textbf{0.997} & \textbf{0.997} & \textbf{0.997} & \textbf{0.997} & 0.996 & 0.996 \\
\midrule \multicolumn{9}{l}{\emph{MLOmics}} \\
BRCA CNV & 0.912 & \textbf{0.920} & 0.909 & 0.918 & 0.914 & 0.919 & 0.885 & 0.907 \\
BRCA methylation & 0.949 & \textbf{0.952} & 0.947 & 0.950 & 0.937 & 0.948 & 0.942 & \textbf{0.952} \\
BRCA miRNA & 0.958 & \textbf{0.961} & 0.957 & \textbf{0.961} & 0.954 & 0.960 & 0.950 & 0.954 \\
BRCA mRNA & 0.982 & \textbf{0.983} & 0.981 & \textbf{0.983} & 0.976 & 0.981 & 0.969 & 0.980 \\
BRCA all omics & 0.983 & \textbf{0.984} & 0.981 & 0.983 & 0.974 & 0.980 & 0.965 & 0.977 \\
COAD CNV & 0.896 & 0.903 & 0.902 & 0.901 & 0.900 & 0.903 & 0.901 & 0.904 \\
COAD methylation & 0.917 & \textbf{0.924} & 0.906 & 0.914 & 0.851 & 0.921 & 0.893 & 0.897 \\
COAD miRNA & 0.875 & 0.881 & 0.869 & 0.873 & 0.882 & \textbf{0.906} & 0.857 & 0.877 \\
COAD mRNA & 0.949 & 0.954 & 0.949 & 0.951 & 0.936 & 0.951 & 0.943 & \textbf{0.955} \\
COAD all omics & 0.949 & 0.952 & 0.954 & \textbf{0.958} & 0.922 & 0.940 & 0.912 & 0.936 \\
GBM CNV & 0.809 & \textbf{0.817} & 0.802 & 0.809 & 0.799 & 0.814 & 0.760 & 0.796 \\
GBM methylation & 0.907 & 0.909 & 0.905 & \textbf{0.910} & 0.896 & 0.909 & 0.900 & 0.907 \\
GBM miRNA & 0.902 & \textbf{0.904} & 0.893 & 0.900 & 0.900 & 0.902 & 0.885 & 0.888 \\
GBM mRNA & 0.972 & 0.972 & \textbf{0.973} & \textbf{0.973} & 0.963 & \textbf{0.973} & 0.958 & 0.969 \\
GBM all omics & 0.963 & \textbf{0.964} & \textbf{0.964} & 0.962 & 0.951 & 0.962 & 0.908 & 0.951 \\
LGG CNV & 0.974 & 0.976 & 0.975 & 0.976 & 0.986 & 0.985 & 0.971 & 0.973 \\
LGG miRNA & 0.982 & 0.983 & 0.984 & 0.984 & 0.982 & 0.984 & 0.983 & \textbf{0.985} \\
LGG mRNA & 0.987 & 0.986 & 0.986 & 0.987 & 0.989 & 0.988 & 0.987 & 0.988 \\
OV CNV & 0.793 & \textbf{0.804} & 0.799 & 0.801 & 0.782 & \textbf{0.804} & 0.751 & 0.785 \\
OV methylation & \textbf{0.945} & 0.943 & 0.941 & 0.941 & 0.922 & 0.936 & 0.894 & 0.934 \\
OV miRNA & 0.922 & 0.923 & 0.919 & 0.920 & 0.926 & 0.925 & 0.896 & 0.902 \\
OV mRNA & 0.993 & \textbf{0.994} & 0.992 & 0.993 & 0.988 & 0.993 & 0.978 & 0.989 \\
OV all omics & 0.993 & \textbf{0.994} & 0.992 & 0.993 & 0.979 & 0.986 & 0.951 & 0.981 \\
\midrule \multicolumn{9}{l}{\emph{Shamir}} \\
breast methylation & \textbf{0.961} & \textbf{0.961} & 0.959 & 0.959 & 0.951 & 0.956 & 0.958 & 0.956 \\
breast miRNA & 0.970 & \textbf{0.972} & 0.968 & 0.971 & 0.968 & \textbf{0.972} & 0.967 & 0.970 \\
breast mRNA & \textbf{0.990} & \textbf{0.990} & 0.989 & \textbf{0.990} & 0.982 & 0.988 & 0.988 & \textbf{0.990} \\
breast all omics & \textbf{0.989} & \textbf{0.989} & 0.987 & 0.988 & 0.983 & 0.985 & 0.987 & \textbf{0.989} \\
gbm methylation & 0.894 & \textbf{0.895} & \textbf{0.895} & 0.893 & 0.889 & 0.891 & 0.877 & 0.891 \\
gbm miRNA & 0.903 & 0.908 & 0.905 & 0.908 & 0.880 & 0.898 & 0.876 & 0.886 \\
gbm mRNA & 0.968 & 0.971 & 0.970 & \textbf{0.972} & 0.958 & 0.970 & 0.955 & 0.968 \\
gbm all omics & 0.969 & \textbf{0.972} & 0.967 & 0.971 & 0.949 & \textbf{0.972} & 0.941 & 0.970 \\
sarcoma methylation & \textbf{0.922} & 0.918 & 0.917 & 0.918 & 0.887 & 0.910 & 0.895 & 0.913 \\
sarcoma miRNA & 0.927 & 0.925 & 0.928 & 0.922 & 0.928 & \textbf{0.937} & 0.909 & 0.923 \\
sarcoma mRNA & \textbf{0.935} & 0.934 & 0.931 & \textbf{0.935} & 0.919 & 0.932 & 0.925 & 0.932 \\
sarcoma all omics & 0.935 & \textbf{0.936} & 0.932 & 0.934 & 0.898 & 0.924 & 0.920 & 0.930 \\
\bottomrule
\end{tabular}

%% file: tables/per_task_2.tex
\begin{tabular}{lrrrrrrr}
\toprule
Task & \multicolumn{2}{c}{TabPFNv2} & \multicolumn{2}{c}{TabICLv2} & \multicolumn{1}{c}{XGBoost} & \multicolumn{1}{c}{Random forest} & \multicolumn{1}{c}{Elastic-net LR} \\
\cmidrule(lr){2-3} \cmidrule(lr){4-5} \cmidrule(lr){6-6} \cmidrule(lr){7-7} \cmidrule(lr){8-8}
 & $\times$1 & $\times$8 & $\times$1 & $\times$8 &  &  &  \\
\midrule \multicolumn{8}{l}{\emph{HDLSS}} \\
ALLAML & \textbf{0.998} & 0.997 & 0.955 & 0.991 & 0.963 & 0.995 & 0.993 \\
BASEHOCK & \textbf{0.999} & \textbf{0.999} & 0.991 & 0.997 & 0.991 & 0.996 & 0.997 \\
CLL\_SUB\_111 & 0.778 & 0.896 & 0.721 & 0.774 & 0.873 & 0.854 & 0.885 \\
GLI\_85 & 0.921 & 0.934 & 0.928 & 0.916 & 0.901 & 0.938 & 0.934 \\
PCMAC & 0.979 & 0.981 & 0.966 & 0.971 & 0.960 & 0.976 & 0.955 \\
Prostate\_GE & 0.968 & 0.966 & 0.855 & 0.854 & 0.943 & 0.928 & 0.953 \\
SMK\_CAN\_187 & 0.761 & 0.768 & 0.724 & 0.745 & 0.731 & 0.727 & 0.748 \\
TOX\_171 & 0.943 & 0.982 & 0.939 & 0.965 & 0.925 & 0.943 & 0.983 \\
colon & 0.840 & 0.836 & 0.843 & 0.845 & 0.829 & 0.854 & 0.861 \\
leukemia & 0.997 & 0.998 & 0.989 & 0.981 & 0.992 & 0.995 & 0.992 \\
lung & 0.991 & 0.996 & \textbf{0.997} & \textbf{0.997} & 0.985 & 0.995 & 0.994 \\
\midrule \multicolumn{8}{l}{\emph{MLOmics}} \\
BRCA CNV & 0.861 & 0.902 & 0.911 & 0.917 & 0.884 & 0.876 & 0.836 \\
BRCA methylation & 0.883 & 0.940 & 0.937 & 0.933 & 0.931 & 0.919 & 0.946 \\
BRCA miRNA & 0.952 & 0.956 & 0.956 & \textbf{0.961} & 0.931 & 0.933 & 0.942 \\
BRCA mRNA & 0.937 & 0.972 & 0.966 & 0.967 & 0.966 & 0.967 & 0.965 \\
BRCA all omics & 0.847 & 0.974 & 0.954 & 0.958 & 0.964 & 0.965 & 0.967 \\
COAD CNV & 0.720 & 0.652 & 0.890 & \textbf{0.918} & 0.824 & 0.883 & 0.832 \\
COAD methylation & 0.876 & 0.898 & 0.907 & 0.862 & 0.877 & 0.808 & 0.921 \\
COAD miRNA & 0.861 & 0.879 & 0.893 & 0.902 & 0.843 & 0.777 & 0.859 \\
COAD mRNA & 0.908 & \textbf{0.955} & 0.912 & 0.919 & 0.923 & 0.897 & 0.934 \\
COAD all omics & 0.717 & 0.848 & 0.935 & 0.943 & 0.933 & 0.944 & 0.921 \\
GBM CNV & 0.752 & 0.784 & 0.778 & 0.783 & 0.770 & 0.773 & 0.500 \\
GBM methylation & 0.864 & 0.890 & 0.868 & 0.870 & 0.867 & 0.879 & 0.893 \\
GBM miRNA & 0.874 & 0.880 & 0.903 & 0.901 & 0.859 & 0.867 & 0.891 \\
GBM mRNA & 0.912 & 0.958 & 0.963 & 0.966 & 0.949 & 0.957 & 0.970 \\
GBM all omics & 0.900 & 0.956 & 0.899 & 0.910 & 0.945 & 0.939 & \textbf{0.964} \\
LGG CNV & 0.923 & 0.943 & 0.975 & 0.982 & 0.973 & 0.968 & \textbf{0.987} \\
LGG miRNA & 0.981 & 0.984 & 0.984 & 0.984 & 0.976 & 0.980 & 0.982 \\
LGG mRNA & 0.976 & \textbf{0.991} & 0.984 & 0.984 & 0.987 & 0.987 & 0.987 \\
OV CNV & 0.712 & 0.744 & 0.760 & 0.792 & 0.737 & 0.759 & 0.500 \\
OV methylation & 0.888 & 0.928 & 0.903 & 0.905 & 0.924 & 0.901 & 0.925 \\
OV miRNA & 0.905 & 0.919 & 0.926 & \textbf{0.931} & 0.896 & 0.886 & 0.890 \\
OV mRNA & 0.958 & 0.986 & 0.982 & 0.982 & 0.972 & 0.973 & 0.984 \\
OV all omics & 0.936 & 0.971 & 0.941 & 0.939 & 0.970 & 0.965 & 0.976 \\
\midrule \multicolumn{8}{l}{\emph{Shamir}} \\
breast methylation & 0.956 & 0.948 & 0.944 & 0.947 & 0.947 & 0.941 & 0.950 \\
breast miRNA & 0.958 & 0.971 & 0.967 & 0.970 & 0.947 & 0.939 & 0.906 \\
breast mRNA & 0.983 & 0.985 & 0.978 & 0.979 & 0.978 & 0.982 & 0.970 \\
breast all omics & 0.980 & 0.981 & 0.977 & 0.975 & 0.978 & 0.977 & 0.971 \\
gbm methylation & 0.840 & 0.873 & 0.885 & 0.879 & 0.868 & 0.860 & 0.880 \\
gbm miRNA & 0.864 & 0.900 & 0.895 & 0.903 & 0.890 & 0.873 & \textbf{0.914} \\
gbm mRNA & 0.909 & 0.949 & 0.969 & 0.967 & 0.960 & 0.944 & 0.954 \\
gbm all omics & 0.903 & 0.942 & 0.967 & 0.960 & 0.963 & 0.942 & 0.964 \\
sarcoma methylation & 0.865 & 0.904 & 0.901 & 0.899 & 0.868 & 0.898 & 0.908 \\
sarcoma miRNA & 0.890 & 0.921 & 0.918 & 0.922 & 0.897 & 0.871 & 0.830 \\
sarcoma mRNA & 0.915 & 0.925 & 0.906 & 0.910 & 0.902 & 0.897 & 0.861 \\
sarcoma all omics & 0.916 & 0.923 & 0.907 & 0.912 & 0.908 & 0.921 & 0.881 \\
\bottomrule
\end{tabular}

%% file: tables/development.tex
\begin{tabular}{lrrrrrrr}
\toprule
 & \multicolumn{5}{c}{AUROC} & \multicolumn{2}{c}{TabOmics v2.5 $\times$1 minus model} \\
\cmidrule(lr){2-6} \cmidrule(lr){7-8}
Model & madelon & arcene & RELATHE & gisette & Mean & Development & Reported \\
\midrule
TabPFN-3.5 $\times$8 & \textbf{0.970} & \textbf{0.959} & 0.962 & \textbf{0.999} & \textbf{0.972} & $-0.009$ & $-0.001$ \\
TabPFN-Wide $\times$8 & 0.954 & 0.952 & 0.957 & 0.996 & 0.965 & $-0.005$ & $+0.003$ \\
TabPFN-3.5 $\times$1 & 0.968 & 0.943 & 0.937 & 0.998 & 0.962 & $-0.001$ & $+0.009$ \\
TabPFN-Wide $\times$1 & 0.939 & 0.949 & 0.959 & 0.996 & 0.960 & $-0.003$ & $+0.013$ \\
TabPFNv2 $\times$8 & 0.961 & 0.913 & 0.958 & 0.998 & 0.957 & $+0.001$ & $+0.009$ \\
TabPFNv2 $\times$1 & 0.956 & 0.900 & 0.952 & 0.997 & 0.951 & $+0.004$ & $+0.038$ \\
TabOmics v2.5 $\times$8 & 0.898 & 0.946 & \textbf{0.964} & 0.997 & 0.951 & $-0.003$ & $-0.001$ \\
TabOmics v2 $\times$8 & 0.894 & 0.947 & 0.956 & 0.996 & 0.948 & $-0.002$ & $0.000$ \\
TabOmics v2 $\times$1 & 0.897 & 0.947 & 0.948 & 0.996 & 0.947 & $-0.002$ & $+0.001$ \\
TabOmics v2.5 $\times$1 & 0.886 & 0.942 & 0.961 & 0.997 & 0.947 & -- & -- \\
TabICLv2 $\times$1 & 0.964 & 0.899 & 0.943 & 0.978 & 0.946 & $+0.019$ & $+0.012$ \\
TabICLv2 $\times$8 & 0.967 & 0.899 & 0.940 & 0.977 & 0.946 & $+0.021$ & $+0.012$ \\
XGBoost & 0.852 & 0.870 & 0.937 & 0.996 & 0.914 & $+0.029$ & $+0.024$ \\
Random forest & 0.755 & 0.896 & 0.956 & 0.995 & 0.901 & $+0.026$ & $+0.025$ \\
Elastic-net LR & 0.639 & 0.877 & 0.956 & 0.996 & 0.867 & $+0.035$ & $+0.014$ \\
\bottomrule
\end{tabular}

%% file: tables/top_genes.tex
\begin{tabular}{lrlr}
\toprule
\multicolumn{2}{c}{MLOmics BRCA} & \multicolumn{2}{c}{Shamir breast} \\
\cmidrule(lr){1-2} \cmidrule(lr){3-4}
Gene & Folds & Gene & Folds \\
\midrule
\textbf{FOXC1} & 5 & \textbf{MIA} & 4 \\
GABRP & 5 & KLK8 & 4 \\
SRSF12 & 4 & PPP1R14C & 4 \\
\textbf{KRT5} & 4 & STAC2 & 3 \\
\textbf{FOXA1} & 3 & \textbf{NDC80} & 3 \\
\textbf{SFRP1} & 3 & \textbf{KRT5} & 3 \\
\textbf{ESR1} & 3 & \textbf{CEP55} & 3 \\
\textbf{KRT17} & 3 & \textbf{ESR1} & 3 \\
TRIM29 & 2 & SOX10 & 2 \\
MPHOSPH6 & 2 & \textbf{KRT14} & 2 \\
\bottomrule
\end{tabular}